\documentclass[sn-mathphys,Numbered]{sn-jnl}

\usepackage{graphicx}%
\usepackage{multirow}%
\usepackage{amsmath,amssymb,amsfonts}%
\usepackage{amsthm}%
\usepackage{mathrsfs}%
\usepackage[title]{appendix}%
\usepackage{xcolor}%
\usepackage{textcomp}%
\usepackage{manyfoot}%
\usepackage{booktabs}%
\usepackage[ruled,vlined]{algorithm2e}
\usepackage{algorithmicx}%
\usepackage{algpseudocode}%
\usepackage{listings}%
\usepackage{natbib}
\usepackage[inline]{enumitem}
\setcitestyle{authoryear,open={(},close={)}} 

\theoremstyle{thmstyleone}%
\newtheorem{theorem}{Theorem}
\theoremstyle{thmstyletwo}%

\theoremstyle{thmstylethree}%

\usepackage{amsmath}
\usepackage{amssymb}
\usepackage{amsfonts}
\usepackage{float}
\usepackage{multirow}
\usepackage{amsthm}
\usepackage{soul,xcolor}
\newtheorem{lemma}{Lemma}

\theoremstyle{definition}
\newtheorem{Assumption}{Assumption}

\def \mcH{\mathcal{H}}
\def \mcX{\mathcal{X}}
\def \mcZ{\mathcal{Z}}

\def \mbI{\mathbb{I}}
\def \mcR{\mathcal{R}}
\def \mcB{\mathcal{B}}

\setstcolor{red}

\def \mcA{\mathcal{A}}
\def \bfx{\mathbf{x}}

\def \bfX{\mathbf{X}}
\def \bfx{\mathbf{x}}

\def \bfY{\mathbf{Y}}

\def \ev{\mathbb{E}}
\def \pr{\mathbb{P}}

\newcommand{\vo}{\vec{o}\@ifnextchar{^}{\,}{}}

\newcommand{\floor}[1]{\left\lfloor #1 \right\rfloor}

\usepackage{xparse}
\NewDocumentCommand{\ceil}{s O{} m}{%
    {#2\lceil#3#2\rceil} 
}

\graphicspath{{figure/}}
\begin{document}

\title[Article Title]{Distributed Adaptive Nearest Neighbor Classifier: Algorithm and Theory}


\author*[1]{\fnm{Ruiqi} \sur{Liu}}\email{ruiqliu@ttu.edu}

\author[2]{\fnm{Ganggang} \sur{Xu}}\email{gangxu@bus.miami.edu}

\author[3]{\fnm{Zuofeng} \sur{Shang}}\email{zshang@njit.edu}

\affil*[1]{\orgdiv{Department of Mathematics and Statistics}, \orgname{Texas Tech University}, \orgaddress{\city{Lubbock}, \postcode{79409}, \state{TX}, \country{USA}}}

\affil[2]{\orgdiv{Department of Management Science}, \orgname{University of Miami}, \orgaddress{\city{Coral Gables}, \postcode{33146}, \state{FL}, \country{USA}}}

\affil[3]{\orgdiv{Department of Mathematical Sciences}, \orgname{New Jersey Institute of Technology}, \orgaddress{\city{City}, \postcode{07102}, \state{NJ}, \country{USA}}}


\abstract{When data is of an extraordinarily large size or physically stored in different locations, the distributed nearest neighbor (NN) classifier is an attractive tool for classification. We propose a novel distributed adaptive NN classifier for which the number of nearest neighbors is a tuning parameter stochastically chosen by a data-driven criterion. An early stopping rule is proposed when searching for the optimal tuning parameter, which not only speeds up the computation but also improves the finite sample performance of the proposed algorithm. Convergence rate of excess risk of the distributed adaptive NN classifier is investigated under various sub-sample size compositions. In particular, we show that when the sub-sample sizes are sufficiently large, the proposed classifier achieves the nearly optimal convergence rate. Effectiveness of the proposed approach is demonstrated through simulation studies as well as an empirical application to a real-world dataset.}

\keywords{Distributed Learning, Adaptive Procedure, Minimax Optimal, Binary Classification}



\maketitle

\section{Introduction}\label{Sec:introduction}
Nearest neighbor (NN) classifier is a simple but powerful tool for various applications such as text classification (\citealp{han2001text,jiang2012improved}), query dependent ranking (\citealp{geng2008query}), and pattern recognition (\citealp{kowalski1972k,zheng2004locally,xu2013improvement}). Consider $(Y_1, \bfX_1),\ldots, (Y_N, \bfX_N)$ generated independently from an unknown probability distribution $P$, with $Y_i\in \{0, 1\}$ being the label and $\bfX_i$ being the corresponding $d$-dimensional feature vector for $i=1,\cdots,N$. The NN classifier predicts the label of a query point $\bfx$ based on labels of its neighboring observations. It is well-known that NN algorithm is sensitive to the scale of data as it relies on computing the distances. A popular procedure is to normalize each feature to $[0, 1]$. Without loss of generality, we assume that the feature space is $[0,1]^d$ and that the Euclidean distance is used. This assumption was also used in \cite{cai2019transfer}. Given a  new query point $\bfx\in [0, 1]^d$,  denote $\bfX_{(i)}(\bfx)$ as the $i$-th nearest  point to $\bfx$ among $\bfX_1,\ldots, \bfX_N$, and $Y_{(i)}(\bfx)$ as the label associated with $\bfX_{(i)}(\bfx)$. For a prespecified integer $1\leq k\leq N$, the conditional probability $\eta(\bfx):=\pr(Y=1|\bfX=\bfx)$ can be approximated by the $k$-NN estimator $\widehat{\eta}_{NN,k}(\bfx)=\frac{1}{k}\sum_{i=1}^kY_{(i)}(\bfx)$ and the label associated with $\bfx$ is then predicted as $\widehat{f}_{NN,k}(\bfx)=\mbI(\widehat{\eta}_{NN}(\bfx)\geq 1/2)$, with $\mbI(\cdot)$ being the indicator function.

The performance of a binary classifier $\widehat{f}: [0, 1]^d\to \{0, 1\}$, which is trained using observed data  $(Y_1, \bfX_1),\ldots, (Y_N, \bfX_N)$, is commonly evaluated by the regret (or excess risk) defined as
\begin{eqnarray}
\mcR(\widehat{f})=\pr(\widehat{f}(\bfX)\neq \bfY)-\pr(f^*(\bfX)\neq \bfY),\nonumber
\end{eqnarray}
where $(Y, \bfX)\sim P$ is an independent copy of the training sample, $f^*(\bfx)=\mbI(\eta(\bfx)\geq 1/2)$ is the well-known Bayesian classifier, and the probability is with respect to the joint distribution of  $(Y_1, \bfX_1)$, $\ldots, (Y_N, \bfX_N)$ and $(Y, \bfX)$. A smaller regret indicates  higher classification accuracy for a classifier $\widehat{f}$.

\noindent \textbf{Notation:} For deterministic positive sequences $a_N$ and $b_N$, we denote $a_N\lesssim (\textrm{or} \gtrsim)\; b_N$ if $a_N\leq (\textrm{or} \geq)\; Cb_N$ for some $C>0$ and sufficiently large $N$. If $a_N\lesssim b_N$ and $a_N\gtrsim b_N$, we write $a_N\asymp b_N$. For any $a>0$, we denote $\ceil{a}$ ($\floor{a}$) as the smallest (largest) integer that is not less (greater) than $a$. We denote $\lambda$ as the Lebesgue measure and $P_\bfX$ as the  marginal distribution of $\bfX$ whose support is $\Omega$. For a set $A$, we use $|A|$ to denote its cardinality.

\subsection{Related Work}
The regret of the $k$-NN classifier has been shown to converge to $0$ as $k \to \infty$ and  $k/N \to 0$ in a general metric space with additional structural assumptions (\citealp{cover1967nearest, cerou06knninfinite, knnconsistency2021}) and in the Euclidean space (\citealp{stone1977consistentnp, devroye1994strong}). The convergence rate of the regret depends on properties of $\eta(\bfx)$ and $P_\bfX$. \cite{chaudhuri2014rates} established a nonasymptotic bound for the convergence rate, {which achieves the minimax rate in the sense of \cite{audibert2007fast} under some mild conditions.} \cite{gadatknn2016} further identified two sufficient and necessary conditions for the uniform consistency of the $k$-NN classifier without rigid assumptions on the joint distribution of $(Y,\bfX)$ and derived the corresponding optimal convergence rate. \cite{samworth2012} proposed an optimally weighted $k$-NN classifier based on a new asymptotic expansion of its regret.

When facing an extraordinarily large sample size, the $k$-NN classifier can be computationally intensive, especially when $k$ is large.  To address this issue, \cite{qiao2019rates} and  \cite{duan2018distributed} proposed two distributed $k$-NN classifiers, extending the work of \cite{chaudhuri2014rates}  and \cite{samworth2012}, respectively. Their algorithms first divide the whole data into $m$ equally-sized sub-samples, and for each sub-sample, a $k$-NN classifier is trained independently. The final prediction of a new query point is made by aggregating the $m$ independently trained $k$-NN classifiers. Under suitable conditions, the regrets of both distributed $k$-NN classifiers were shown to achieve the optimal convergence rate. However, in many applications, the sub-samples may not have equal sample sizes, and to the best of our knowledge, there has yet been any existing work on distributed NN classifiers with unequal sized sub-samples.

Furthermore, the aforementioned theoretical results are based on the key assumption that the choice of $k$ is pre-given and is deterministic. However, it is often desirable to have a data-driven choice of $k$ for practical applications. There has been limited work on theoretical properties of the $k$-NN classifier with a data-driven choice of $k$ in existing literature, with two notable exceptions, i.e.,  \cite{cai2019transfer} and \cite{balsubramani2019adaptive}. They independently proposed two adaptive procedures to stochastically choose $k$ and established the convergence rates of the resulting adaptive NN classifiers under suitable conditions. However, while achieving improved classification accuracy, searching for an optimal $k$ also significantly increases the computational burden for the adaptive NN classifier, making it desirable to consider a distributed adaptive NN classifier  with favorable statistical properties when the sample size $N$ is extraordinarily large. For applications where data are stored in different locations, a distributed adaptive NN classifier is also a natural and preferable choice.


\subsection{Our Contribution}

We propose a novel distributed adaptive NN classifier with a data-driven choice of $k$, which can be used to either speed up the computation when the data size is extraordinary large or improve the classification accuracy when data are stored in different machines. Suppose that the whole data set is separately stored in $m$ different locations, and each location has a sub-sample of size $n_j$, $j=1,\cdots,m$. The sub-sample sizes are allowed to be different from each other, in contrast to the existing divide-and-conquer framework \citep{duan2018distributed, qiao2019rates}.  Without loss of generality, we assume that $n_1\geq n_2\geq \ldots \geq n_m$ and denote $N=n_1+\cdots+n_m$. Based on the $j$th sub-sample, a local $k_j$-NN classifier is constructed for a given query point $\bfx$ and an integer $k_j$, $j=1,\cdots,m$. The predicted label for $\bfx$ is then obtained by aggregating the $m$ sub-sample NN classifiers with $k_1,\cdots,k_m$ chosen by a data-driven criterion. See Section~\ref{section:model} for more details. 

The computational efficiency of the proposed algorithm is achieved in two ways.

 {\it\bf (1) Parallel computation.} For a given $k$, the computational complexity of the standard $k$-NN classifier using the whole data is between $O(N)$ to $O(N\log(N))$ (\citealp{cormen2009introduction}), which needs to be carried out on a single machine. In comparison, the computation of the distributed NN classifier can be easily paralleled, and each sub-sample only costs between $O(n)$ to $O(n\log(n))$ operations.
 
   {\it\bf (2) Early stopping rule for $k$.}  The adaptive NN classifiers proposed in \cite{cai2019transfer} and \cite{balsubramani2019adaptive} search for an optimal $k_1$ by increasing $k$ from $1$ to $N$ until a stopping rule is triggered.  A straightforward extension of their approaches to the distributed setting is to search for $k_j$ from $1$ to $n_j$, $j=1,\cdots,m$.  However, we propose an early stopping rule for the choice of $k_1$ (which determines other $k_j$'s), narrowing down the search range for $k_1$ to $\{1,\ldots,  \ceil{n_1N^{-\frac{d}{2+d}}\log(N)}\}$. As a result, the proposed algorithm significantly reduces the number of attempts needed to locate the optimal $k_j$'s for the distributed adaptive NN classifier. 
   
Our numerical studies show that such an early stopping rule for $k_1$ not only speeds up the computation but also yields superior finite sample performance for the proposed algorithm compared to the naive extension of  \cite{cai2019transfer} and \cite{balsubramani2019adaptive}. See Section~\ref{Sec:simulation} for more details.

From a theoretical point of view, our work extends the theory for distributed NN classifier with a fixed $k$ \citep{qiao2019rates} to the more realistic distributed adaptive NN classifier based on unequal sub-sample sizes, whose $k_j$'s are chosen by a data-driven procedure. Specifically, we derive the convergence rate of the regret of the proposed classifier and give sufficient conditions under which the convergence rate is optimal (up to logarithmic factors). {Moreover, the convergence rate of the regret exhibits a phase transition characterized by sub-sample sizes. } Finally, we wish to comment that  the proof of adaptivity in the distributed framework relies on the uniform convergence in Lemma \ref{lemma:uniform:concentration:eta:hat:m:version}. This requires bounding the total model complexity (see Lemma  \ref{lemma:counting:vc}) of all the local classifiers, which motivates the choices of $k_j$'s in Algorithm \ref{alg:distributed}.


The rest of this paper is structured as follows. Section \ref{section:model} introduces the algorithm for the distributed adaptive NN classifier and Section~\ref{theory} investigates its asymptotic properties. Section \ref{Sec:simulation} carries out a set of simulation studies, and a real-world dataset is analyzed in Section \ref{sec:realdata}.
All technical proofs are provided in the Appendix.

\section{Distributed Adaptive Nearest Neighbor Classifier}\label{section:model}
Suppose that the whole dataset, denoted as $\mcZ=\{(Y_1, \bfX_1),\ldots, (Y_N, \bfX_N)\}$, are distributed across $m$ machines. Each machine hosts a sub-sample of size $n_j$, denoted as $\mcZ_j=\{(Y_1^j, \bfX_1^j),\ldots, (Y_{n_j}^j, \bfX_{n_j}^j)\}$ for $j=1,\cdots,m$. 
For the $j$th sub-sample, given an integer $k_j\in \{1,\ldots, n_j\}$,  the $j$th local NN estimator of $\eta(\bfx)=\pr(Y=1|\bfX=\bfx)$ for a new query point $\bfx \in [0, 1]^d$ is defined as
 \begin{equation*}
\widehat{\eta}_{k_j,j}(\bfx)=\frac{1}{k_j}\sum_{i=1}^{k_j}Y_{(i)}^j(\bfx),\quad j=1,\ldots,m,
\end{equation*}
where $Y_{(i)}^j(\bfx)$ is the label associated with $\bfX_{(i)}^j(\bfx)$, the $i$-th nearest neighbors of $\bfx$ among $\bfX_1^j,\ldots,$ $\bfX_{n_j}^j$. The proposed distributed NN classifier is subsequently defined as 
\begin{eqnarray}
\widehat{f}_{k_1:k_m}(\bfx)=\mbI(\widehat{\eta}_{k_1:k_m}(\bfx)\geq 1/2)\quad  \textrm{ with } \quad \widehat{\eta}_{k_1:k_m}(\bfx)=\frac{1}{\sum_{j=1}^m k_j}\sum_{j=1}^m k_j \widehat{\eta}_{k_j, j}(\bfx),\label{knnuneq}
\end{eqnarray}
where the integer sequence $k_1,\ldots,k_m$ need to be chosen by some data-driven method.

The performance of the classifier \eqref{knnuneq} depends critically on the choice of  $k_1,\ldots,k_m$.   The following Algorithm~\ref{alg:distributed} is designed in the same spirit of \cite{cai2019transfer} and \cite{balsubramani2019adaptive}.

\begin{algorithm}[h!]
\KwIn{new query $\bfx$, {training samples $\mcZ_j$,} $j=1,\ldots, m$;}
\textbf{Initialization: } set $k_1=0$;

\textbf{while} $k_1\leq  n_1N^{-\frac{d}{2+d}}\log(N)$ \textbf{do}

\hspace*{0.5cm}  update $k_1:=k_1+1$;

\hspace*{0.5cm}  update $k_j:=\ceil{k_1n_j/n_1}$ and calculate $\widehat{\eta}_{k_j, j}(\bfx)$ for $j=1,\ldots, m$;

\hspace*{0.5cm} calculate $\widehat{\eta}_{k_1:k_m}(\bfx)$ and $r_{k_1}=\sqrt{2\sum_{j=1}^m k_j}|\widehat{\eta}_{k_1:k_m}(\bfx)-1/2|$;

\hspace*{0.5cm} \textbf{if } $r_{k_1}>\sqrt{(d+2)\log(N)}$  \textbf{or} $k_1\geq n_1N^{-\frac{d}{2+d}}\log(N)$ \textbf{ then} 

\hspace*{1cm}   set $\widehat{k}_1=k_1$ and $\widehat{k}_j=\ceil{\widehat{k}_1n_j/n_1}$ for $j=2,\ldots, m$;

\hspace*{1cm}  calculate $\widehat{\eta}_{\widehat{k}_1:\widehat{k}_m}(\bfx)$;

\hspace*{1cm}  exit  loop;

\hspace*{0.5cm} \textbf{end if}

\textbf{end while}

\KwOut{classifier $\widehat{f}_{\widehat{k}_1:\widehat{k}_m}(\bfx)=\mbI(\widehat{\eta}_{\widehat{k}_1:\widehat{k}_m}(\bfx)\geq 1/2)$.}
 \caption{Distributed Adaptive NN Classifier}
 \label{alg:distributed}
\end{algorithm}

 Algorithm~\ref{alg:distributed} assumes that each $k_j$ is proportional to $n_j$ for $j=1,\cdots,m$, and search for the optimal $k_1$ within the set $\{1, \ldots, \ceil{n_1N^{-\frac{d}{2+d}}\log(N)} \}$ such that $|\widehat{\eta}_{k_1:k_m}(\bfx)-1/2|$ based on the classifier~\eqref{knnuneq}
 is strictly greater than $\sqrt{(d+2)\log(N)/(2\sum_{j=1}^mk_j)}$. If no $k_1$ meets this criterion, we simply set $k_1=\ceil{n_1N^{-\frac{d}{2+d}}\log(N)}$. We comment that a naive extension of~\cite{cai2019transfer} and \cite{balsubramani2019adaptive} to the distributed data setting would require searching for $k_1$ from $1$ to $n_1$. In this sense, the upper bound $\ceil{n_1N^{-\frac{d}{2+d}}\log(N)}$ in Algorithm~\ref{alg:distributed} serves an early stopping rule for the search of $k_1$. Our simulation studies demonstrate that such an early stopping rule yields superior finite sample performance compared to the same algorithm but searches $k_1$ from $1$ to $n_1$.
 
An intuitive justification of  Algorithm~\ref{alg:distributed} is as follows. Denote $\mcX=\{\bfX_1,\cdots,\bfX_N\}$.  Under suitable conditions, one can show that $|\widehat{\eta}_{k_1:k_m}(\bfx)-\ev(\widehat{\eta}_{k_1:k_m}(\bfx)|\mcX)|$ is bounded by the sequence $\sqrt{(d+2)\log(N)/(2\sum_{j=1}^mk_j)}$ {uniformly for all $\bfx$ and $k_1,\ldots, k_m$} with a high probability. The stopping rule designed in Algorithm~\ref{alg:distributed} thus ensures that  $\widehat{\eta}_{\hat k_1:\hat k_m}(\bfx)-1/2$ and $\ev(\widehat{\eta}_{\hat k_1:\hat k_m}(\bfx)|\mcX)-1/2$ have the same sign with a high probability. Under suitable conditions, $\ev(\widehat{\eta}_{\hat k_1:\hat k_m}(\bfx)|\mcX)$ is a consistent estimator of $\eta(\bfx)$, which further implies that the distributed adaptive NN classifier $\widehat{f}_{\hat k_1:\hat k_m}(\bfx)=\mbI(\widehat{\eta}_{\hat k_1:\hat k_m}(\bfx)\geq 1/2)$ is asymptotically equivalent to the Bayesian classifier $f^*(\bfx)=\mbI(\eta(\bfx)\geq 1/2)$.

\section{Asymptotic Properties}
\label{theory}
\subsection{Technical Assumptions }
To investigate the asymptotic properties of the proposed adaptive distributed NN classifier obtained from  Algorithm~\ref{alg:distributed}, several technical assumptions are needed.

\begin{Assumption}\label{A1:strong:density} (Strong Density)  For some constants $c_\lambda, r_\lambda>0$, it holds that
(a) \label{A1:strong:density:b} $\lambda [\Omega \cap B(\bfx, r)]\geq c_\lambda \lambda[B(\bfx,r)]$ for all $0<r<r_\lambda$ and $\bfx\in \Omega$; and
(b)\label{A1:strong:density:c}  $c_\lambda<\frac{dP_{\bfX}}{d\lambda}(\bfx)<c_\lambda^{-1}$  for all $\bfx \in \Omega$. 

\end{Assumption}
\begin{Assumption}\label{A1:holder:smooth} (Smoothness) There exist constants $\beta\in (0, 1]$ and $C_\beta>0$ such that $|\eta(\bfx_1)-\eta(\bfx_2)|\leq C_\beta \|\bfx_1-\bfx_2\|^{\beta}$ holds for all $\bfx_1, \bfx_2 \in \Omega$.
\end{Assumption}
\begin{Assumption}\label{A1:marginal:assumption}(Marginal Assumption) For some constants $\alpha\in [0, d/\beta]$ and $C_\alpha>0$ and  all $t\in (0, 1/2]$, the inequality $\pr(|\eta(\bfX)-1/2|<t)\leq C_\alpha t^{\alpha}$ holds.
\end{Assumption}

Assumption \ref{A1:strong:density} is the so-called strong density assumption \citep{audibert2007fast} that imposes two conditions on the  distribution of the feature vector $\bfX$. In particular, \ref{A1:strong:density}(a) requires that the support $\Omega$ does not contain any isolate points and \ref{A1:strong:density:c}(b) assumes the probability density of $\bfX$ is bounded above and below in its support, as commonly required in the literature (e.g., \citealp{h98, h03}). Assumption \ref{A1:holder:smooth} is the uniform Lipschitz condition imposed on the conditional probability $\eta(\bfx)$, { and similar conditions were imposed in \cite{audibert2007fast, gadatknn2016,cai2019transfer}.}  Assumption \ref{A1:marginal:assumption} is a popular condition in classification problems (e.g., see \citealp{audibert2007fast, gadatknn2016}), which characterizes the strength of the signal $|\eta(\bfX)-1/2|$. With a larger $\alpha$, $\eta(\bfX)$ is near the decision boundary $1/2$ with a lower probability, leading to an easier classification problem.

\subsection{Theoretical Results in General Setting}\label{sec:theory1}
In this section, we first present some theoretical results on the distributed adaptive NN classifier in a general setting where sub-sample sizes (i.e., $n_j$'s) are allow to be different. The following theorem gives an upper bound of the regret of the proposed classifier in Algorithm \ref{alg:distributed}.
\begin{theorem}\label{theorem:adaptive:estimator:unequal}
Under Assumptions \ref{A1:strong:density}-\ref{A1:marginal:assumption} and $\min\limits_{1\leq j\leq m}n_j\gtrsim N^{1-\epsilon}$ for some $\epsilon<\frac{2\beta}{2\beta+d}$\;. It follows that
$$\mcR(\widehat{f}_{\widehat{k}_1:\widehat{k}_m})\lesssim  \left[{N}/{\log(N)}\right]^{-\frac{\beta(1+\alpha)}{2\beta+d}}.$$
\end{theorem}
The proof is given in the Appendix.

Theorem \ref{theorem:adaptive:estimator:unequal} establishes the convergence rate of the proposed classifier when sub-sample sizes are not too small, i.e., $\min_j n_j\gtrsim N^{1-\epsilon}$ for some $\epsilon<2\beta/(2\beta+d)$. We remark that this convergence rate coincides with the minimax lower bound  given in \cite{audibert2007fast} up to a logarithm factor. The additional $\log(N)$ term is the price to pay for the adaptive choice of tuning parameters $k_1,\cdots,k_m$, as commonly seen in the literature (e.g., see \citealp{lepskii1991problem, lepski1997optimal}). 

To shed more lights on this issue, we consider a distributed NN classifier with a non-stochastic choice of tuning parameter satisfying $k_j \asymp n_jN^{-\frac{d}{2\beta+d}}$, $j=1,\cdots,m$, which is essentially an extension of \cite{qiao2019rates} which only considered the case $n_1=\cdots=n_m$. The following theorem gives an upper bound of the regret of the resulting distributed NN classifier given in (\ref{knnuneq}).

\begin{theorem}\label{theorem:deterministic:estimator:unequal}
(Non-adaptive $k_j$'s) Suppose that Assumptions \ref{A1:strong:density}-\ref{A1:marginal:assumption} hold and $\min\limits_{1\leq j\leq m}n_j\gtrsim N^{1-\epsilon}$ for some $\epsilon<\frac{2\beta}{2\beta+d}$\;. Then if $k_j \asymp {n_jN^{-\frac{d}{2\beta+d}}}$\; for $j=1,\ldots, m$,  it follows that $$\mcR(\widehat{f}_{k_1:k_m})\lesssim  N^{-\frac{\beta(1+\alpha)}{2\beta+d}}.$$
\end{theorem}
The proof is given in the Appendix.

Theorem~\ref{theorem:deterministic:estimator:unequal} asserts that if $k_j$'s are not chosen by a data-driven method, the minimax lower bound  of the regret \citep{audibert2007fast} is achieved by the distributed NN classifier provided that $k_j=C_j(n_jN^{-\frac{d}{2\beta+d}})$ for some constant $C_j>0$, $j=1,\cdots,m$.  Although Theorem~\ref{theorem:deterministic:estimator:unequal} is of limited practical interest since it is difficult to determine {the values of $C_j$'s and $\beta$} for a given data set,  it indeed motivates us to propose the early stopping threshold $\ceil{n_1N^{-\frac{d}{2\beta+d}}\log(N)}$ when searching for the optimal $k_1$ in Algorithm~\ref{alg:distributed}, which resulted in an extra $\log(N)$ term in its regret convergence rate as suggested by Theorem \ref{theorem:adaptive:estimator:unequal}.

Even though the convergence rates in Theorems  \ref{theorem:adaptive:estimator:unequal} and \ref{theorem:deterministic:estimator:unequal} look similar, their proofs rely on completely different techniques. Since $k_1,\ldots, k_m$ are deterministic in Theorem \ref{theorem:deterministic:estimator:unequal}, the regret of $\widehat{f}_{k_1:k_m}$ can be established through calculating its bias and variance. However, when $\widehat{k}_1,\ldots, \widehat{k}_m$ are data-driven, the regret of $\widehat{f}_{\widehat{k}_1:\widehat{k}_m}$ requires more sophisticated analysis. One major difficulty, for instance, is quantifying the model complexity, which relies on the following lemma.

\begin{lemma}\label{lemma:counting:vc}
Given observations $\{\bfX_1^1,\ldots, \bfX_{n_1}^1,\ldots, \bfX_{1}^m,\ldots, \bfX_{n_m}^m\}$, for $k_j\in \{1,\ldots, n_j\}$ with $j=1,\ldots, m$, we define sets
\begin{eqnarray*}
\mcA_{k_j,j}(\bfx)&:=&\{\bfX_{(1)}^j(\bfx),\ldots, \bfX_{(k_j)}^j(\bfx)\},\\
\mcB&:=&\mcB(k_1, \ldots, k_m)=\{\mcA_{k_1,1}(\bfx)\times \ldots \times \mcA_{k_m,m}(\bfx): \bfx \in [0, 1]^d\}.
\end{eqnarray*}
Then the cardinality of $\mcB$ is bounded by $dN^d$.
\end{lemma}
The proof is given in the Appendix.

Lemma \ref{lemma:counting:vc} counts the number of sets of the form $\mcA_{k_1,1}(\bfx)\times \ldots \times \mcA_{k_m,m}(\bfx)$ when $\bfx$ is running over $[0, 1]^d$. It shows that this number is upper bounded by $dN^d$. This is a generalization of Lemma 3 in \cite{jiang2019non} from $m=1$ to $m>1$. The selection of $k_1, \ldots, k_m$ can be viewed as a model selection problem with $n_1 \times \ldots \times n_m$ candidate models, and the complexity of each model is measured by $|\mcB(k_1, \ldots, k_m)|$. The proof of Theorem \ref{theorem:adaptive:estimator:unequal} requires controlling the complexity of all the candidate models. If we do not specify any constraints on the $k_j$'s and allow for all the combinations of $k_1, \ldots, k_m$, then the complexity of all the candidates models can be evaluated by the following:
\begin{eqnarray*}
|\cup_{k_m=1}^{n_m}\ldots \cup_{k_1=1}^{n_1}\mcB(k_1,\ldots, k_m)|\leq dN^d \times n_1\times\ldots \times  n_m,
\end{eqnarray*}
which is relatively large. As a matter of fact, if we impose a restriction that $k_j=\ceil{ k_1n_j/n_1}$ for all $j=1,\ldots, m$, then we only need to conduct model selection among $n_1$ models, and the corresponding complexity can be bounded by
\begin{eqnarray*}
|\cup_{k_1=1}^{n_1} \cup_{k_2=\ceil{ k_1n_2/n_1}}\ldots \cup_{k_m=\ceil{ k_1n_m/n_1}} \mcB(k_1,\ldots, k_m)|\leq dN^d \times n_1.
\end{eqnarray*}
This reduced complexity plays an important role in deriving the near optimal rate in Theorem  \ref{theorem:adaptive:estimator:unequal}, and it also  motivates the choice $k_j=\ceil{ k_1n_j/n_1}$ in  Algorithm \ref{alg:distributed}. 

%
\subsection{Theoretical Results with $n_1=\cdots=n_m$}
Theorem \ref{theorem:adaptive:estimator:unequal} is limited to the case when $\min\limits_{1\leq j\leq m}n_j\gtrsim N^{1-\epsilon}$ for some $\epsilon<{2\beta}/{(2\beta+d)}$\;, where it asserts that the optimal convergence rate (up to a factor of $\log(N)$) of the regret can be achieved by the proposed classifier. However, theoretical properties of the proposed classifier are unclear when  $\min\limits_{1\leq j\leq m}n_j\gtrsim N^{1-\epsilon}$ only holds for some $\epsilon\geq 2\beta/(2\beta+d)$. While it is difficult to study in general, we manage to provide a partial answer by considering the special case $n_1=\cdots=n_m$, which has been widely studied under the so-called ``divide-and-conquer" framework~\citep{zhang2015divide, shang2017computational, xu2018optimal, qiao2019rates,shang2019nonparametric,dxu2019ddcv,duan2018distributed}. 

\begin{theorem}\label{theorem:adaptive:estimator}
 Suppose that Assumptions \ref{A1:strong:density}-\ref{A1:marginal:assumption} hold and that $n_1=\cdots=n_m=n\asymp N^{1-\epsilon}$ for some $\epsilon\in[0,1)$\;, then it holds that
(a) if $\epsilon<\frac{2\beta}{2\beta+d}$, then $\mcR(\widehat{f}_{\widehat{k}_1:\widehat{k}_m})\lesssim  \left[{N}/{\log(N)}\right]^{-\frac{\beta(1+\alpha)}{2\beta+d}}$; (b) if $\epsilon\geq  \frac{2\beta}{2\beta+d}$, then   $\mcR(\widehat{f}_{\widehat{k}_1:\widehat{k}_m})\lesssim  [\log(N)]^\Delta\left[{N}/{\log(N)}\right]^{-\frac{(1-\epsilon)\beta(1+\alpha)}{d}}$ for some $\Delta>0$.
\end{theorem}
The proof is given in the Appendix.
%

Theorem \ref{theorem:adaptive:estimator} characterizes the asymptotic behavior of the proposed classifier in two scenarios.  When  $\epsilon<2\beta/(2\beta+d)$, part (a) is a special case of Theorem \ref{theorem:adaptive:estimator:unequal}, where the regret convergence rate is free of $\epsilon$ and is nearly optimal up to a logarithm factor \citep{audibert2007fast}. However, when $\epsilon\geq 2\beta/(2\beta+d)$, each sub-sample has a smaller sample size, and the resulting convergence rate of the regret becomes $[\log(N)]^\Delta\left[{N}/{\log(N)}\right]^{-\frac{(1-\epsilon)\beta(1+\alpha)}{d}}$ for some constant $\Delta>0$, which slows down when $\epsilon$ increases.  In contrast, the convergence rate in part (a) remains the same as $\epsilon$ changes.

It is unclear whether the convergence rate given in part (b) is optimal since existing literature on distributed NN classifier has mainly focused on the case with $\epsilon<{2\beta}/{(2\beta+d)}$ \citep[e.g.][]{qiao2019rates}. However, we can show that the convergence rate in part (b) is closely related to that of the distributed $1$-NN classifier, as given in the following theorem.

\begin{theorem}\label{theorem:1nn}
 Suppose that Assumptions \ref{A1:strong:density}-\ref{A1:marginal:assumption} hold and that $n_1=\cdots=n_m=n\asymp N^{1-\epsilon}$ for some $\epsilon\in[0,1)$\;. Then if $\epsilon\geq {2\beta}/{(2\beta+d)}$ and fixing $k_1=\cdots=k_m=1$, it holds that
$\mcR(\widehat{f}_{k_1:k_m})\lesssim  [\log(N)]^\Delta\left[{N}/{\log(N)}\right]^{-\frac{(1-\epsilon)\beta(1+\alpha)}{d}}$  for some  $\Delta>0.$
\end{theorem}
The proof is given in the Appendix.

Theorem~\ref{theorem:1nn} shows that the distributed 1-NN classifier can achieve the same convergence rate as the proposed adaptive NN classifier when $\epsilon\geq {2\beta}/{(2\beta+d)}$. This makes intuitive sense because when $\epsilon$ is large, the aggregated classifier~\eqref{knnuneq} averages over a large number of NN classifiers built on sub-samples (i.e., $m=N/n\asymp N^{\epsilon}$) and the overall ``variability" of the resulting aggregated NN classifier can be significantly smaller than its prediction ``bias", which is of the same magnitude of individual NN classifiers from sub-samples. Consequently, to improve the prediction accuracy of the aggregated NN classifier, it is desirable to use the 1-NN classifier for each sub-sample, which has the smallest prediction ``bias" among NN classifiers for a given sample size.  

The similarity between Theorem~\ref{theorem:adaptive:estimator} part (b) and  Theorem~\ref{theorem:1nn} suggests that when $\epsilon\geq {2\beta}/{(2\beta+d)}$, the proposed classifier behave similarly to the distributed 1-NN classifier. This conjecture is supported by our simulation studies in Section~\ref{Sec:simulation} not only in the case where $n_1=\cdots=n_m$ but also in the case where sub-sample sizes are not equal. However, the distributed 1-NN classifier performs much worse than the proposed classifier when $\epsilon$ is small.  {One advantage of the proposed classifier is that it can automatically adjust to both scenarios without the knowledge of the true value of $\beta$.}

\section{Numerical Results}
\subsection{Simulation Studies}\label{Sec:simulation}
In this section, we evaluate the finite sample performance of the proposed algorithm. The following marginal distributions of $\bfX$ will be considered. 
\begin{enumerate}[label={(\alph*}),ref={(\alph*})]
\item $\bfX\sim g_1(\bfx)$: $\bfX=(X_1, X_2, X_3)\in [0, 1]^3$ with $X_1=R\cos(\theta_1)\cos(\theta_2)$, $X_1=R\cos(\theta_1)\sin(\theta_2)$, and  $X_1=R\sin(\theta_1)$. Here $\theta_1, \theta_2\sim Unif(0, 2\pi)$, and $R\sim Unif(0, 1)$ are three independent uniform random variables.
\item $\bfX\sim g_2(\bfx)$: $\bfX=(X_1, X_2, X_3)\in [0, 1]^3$ is generated by a similar process as (a) except $R\sim 0.5 Beta(5,1)+0.5Beta(1,6)$ follows a  Beta mixture distribution.
\end{enumerate}
Given $\bfX=\bfx$, the conditional probability function is $\eta(\bfx)=h(\|\bfx\|)$, where
\begin{eqnarray*}
h(z)=\begin{cases}
0.8 & \textrm{ if } 0\leq z\leq 0.3,\\
 -6z+2.6 & \textrm{ if } 0.3< z\leq 0.4,\\
 0.2 &  \textrm{ if } 0.4< z\leq 0.7,\\
2.6z-1.62 &  \textrm{ if } 0.7< z\leq 0.8,\\
0.46 &  \textrm{ if } 0.8< z\leq 1.
\end{cases}
\end{eqnarray*}
The total sample size is set as {$N=60000$,} and the data are randomly divided into $m=\ceil{N^\epsilon}$, $\epsilon=0, 0.1,\ldots,0.8$, sub-samples by the following two approaches:  

\noindent \textbf{I. Equally Splitting:} The $N$ observations are split into $m$ datasets with (roughly) equal sample size.

\noindent \textbf{II. Unequal Splitting:} The $N$ observations are split into $m$ datasets, and the sample sizes $(n_1,\ldots, n_m)$ follow a multinomial distribution with probabilities $(m/s,\ldots, 1/s)$ for $s=(m+1)m/2$.

For comparison purpose, we consider the following classifiers: 

\noindent  \hskip 1em\textbf{DAES:} The proposed distributed adaptive NN classifier in Algorithm \ref{alg:distributed} with an early stopping bound $\ceil{n_1N^{-\frac{d}{2+d}}}$;

\noindent  \hskip 1em\textbf{DA:}  Modified Algorithm \ref{alg:distributed}, where the early stopping bound is replaced by $n_1$;

\noindent \hskip 1em\textbf{DK} The distributed NN classifier~\citep{qiao2019rates} with   $k_j=\ceil{n_jN^{-\frac{d}{2+d}}}$, $j=1,\ldots, m$.  

\noindent \hskip 1em\textbf{D1:} The distributed 1-NN classifier by setting $k_1=\cdots=k_m=1$ in~\eqref{knnuneq}.

\begin{figure}[htp!]
\centering
\includegraphics[width=2.5 in]{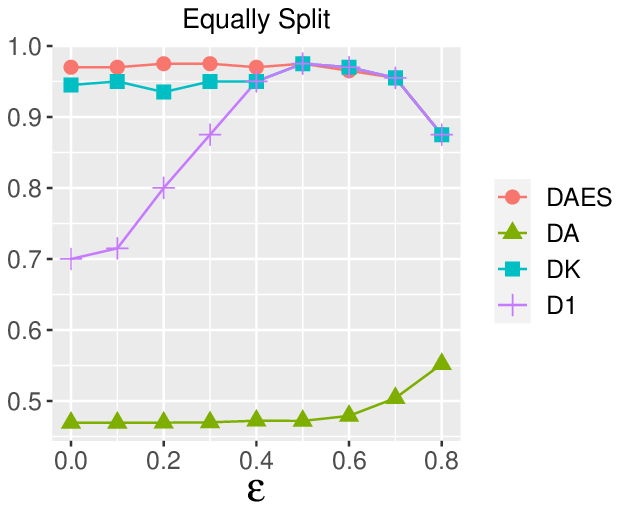}
\includegraphics[width=2.5 in]{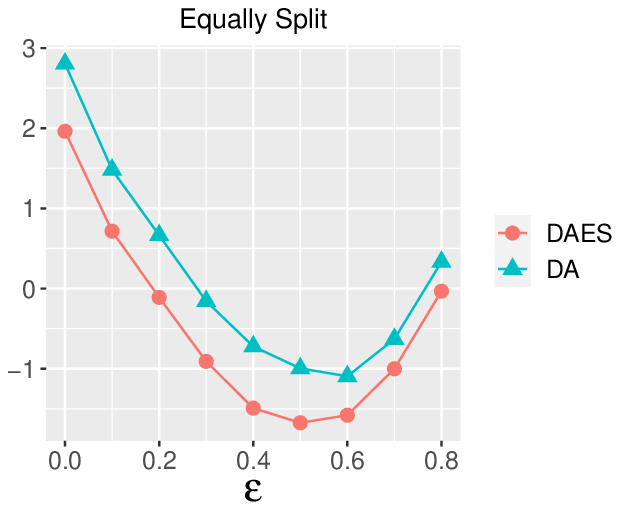}

\includegraphics[width=2.5 in]{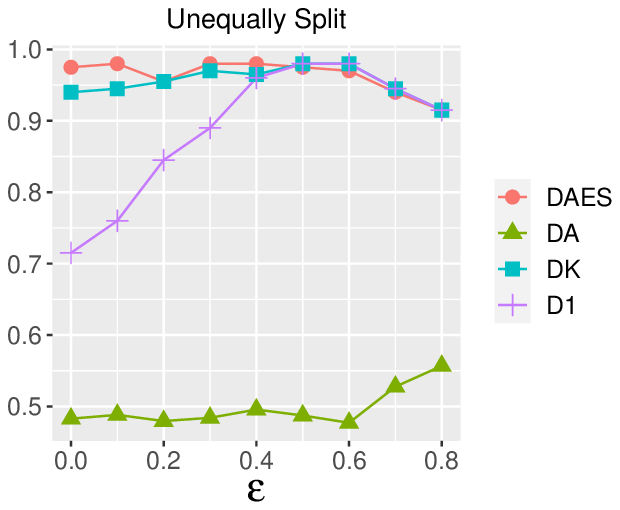}
\includegraphics[width=2.5 in]{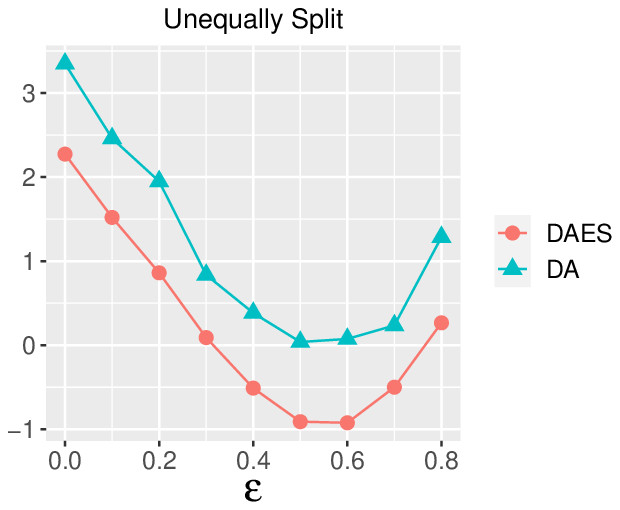}
\caption{Classification accuracy and computation time with $\bfX\sim g_1(\bfx)$. }
\label{figure:risk:time:uniform:x}
\end{figure}

\begin{figure}[htp!]
\centering
\includegraphics[width=2.5 in]{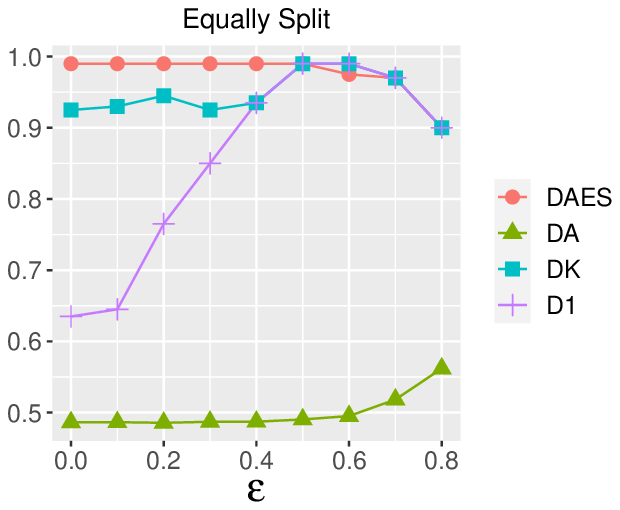}
\includegraphics[width=2.5 in]{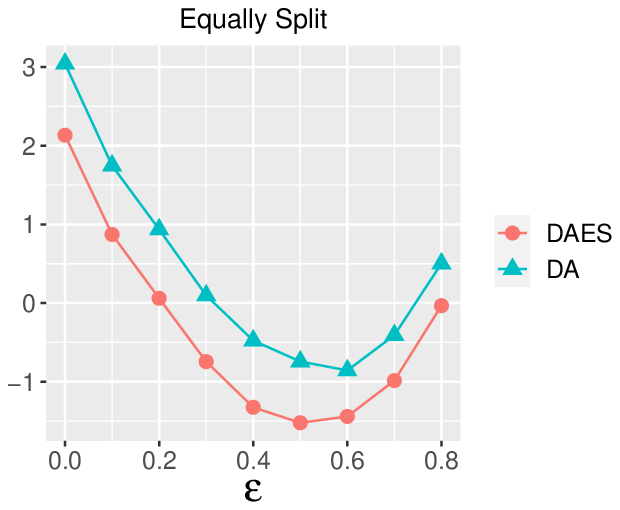}

\includegraphics[width=2.5 in]{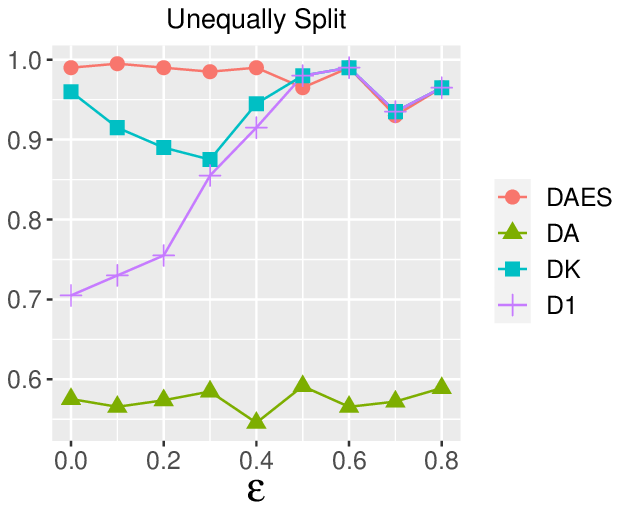}
\includegraphics[width=2.5 in]{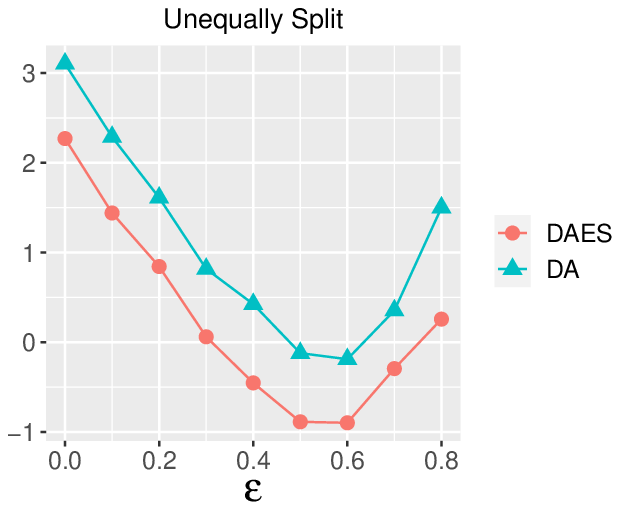}
\caption{Classification accuracy and computation time with $\bfX\sim g_2(\bfx)$.  }
\label{figure:risk:time:beta:x}
\end{figure}

\begin{figure}
\begin{center}
\includegraphics[width=5 in]{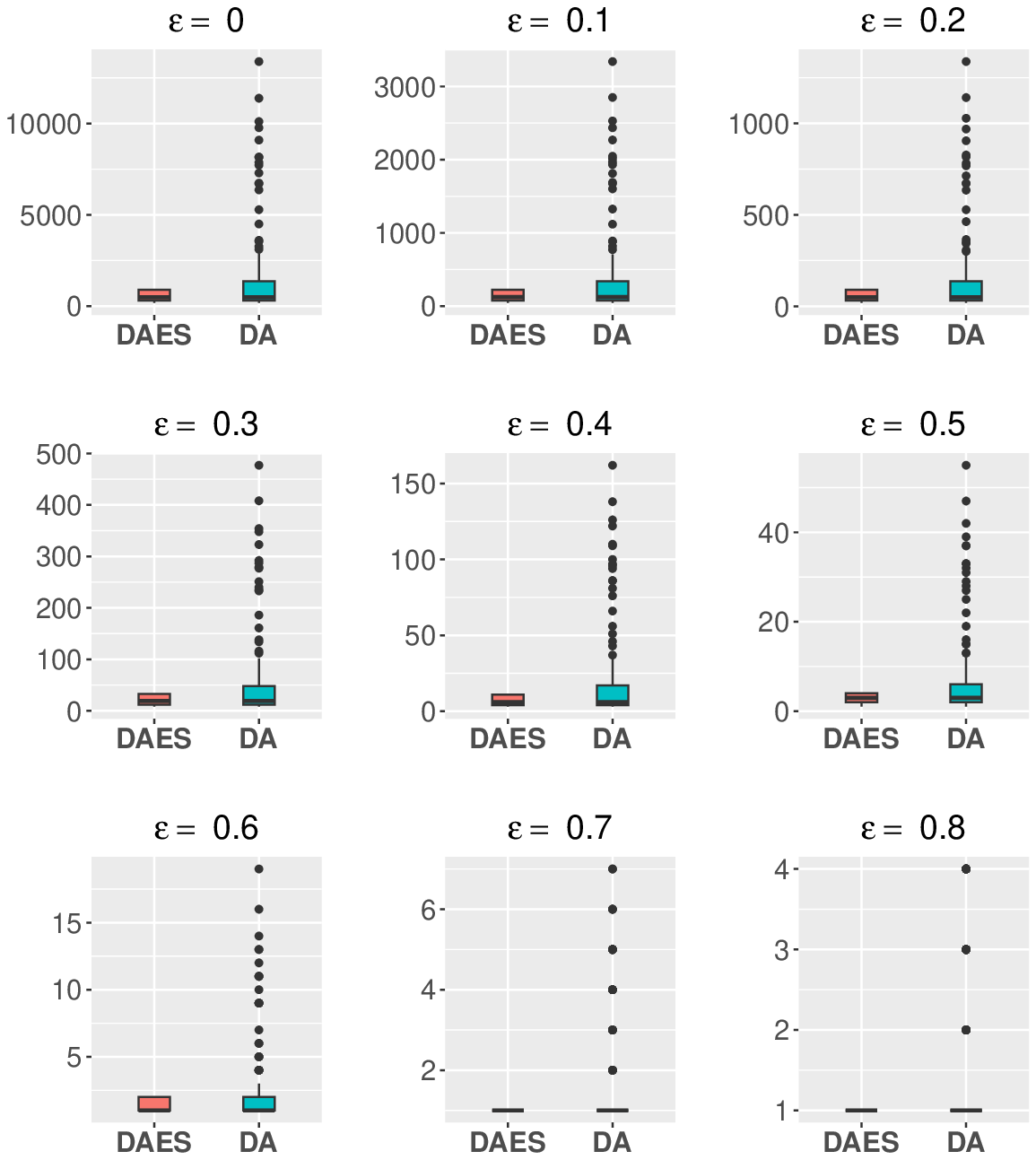}
\end{center}
\caption{Selected $k_1$ with $\bfX\sim g_1(\bfx)$ and equally  split sub-samples.  }
\label{figure:k:uniform:equal:x}
\end{figure}

\begin{figure}
\begin{center}
\includegraphics[width=5 in]{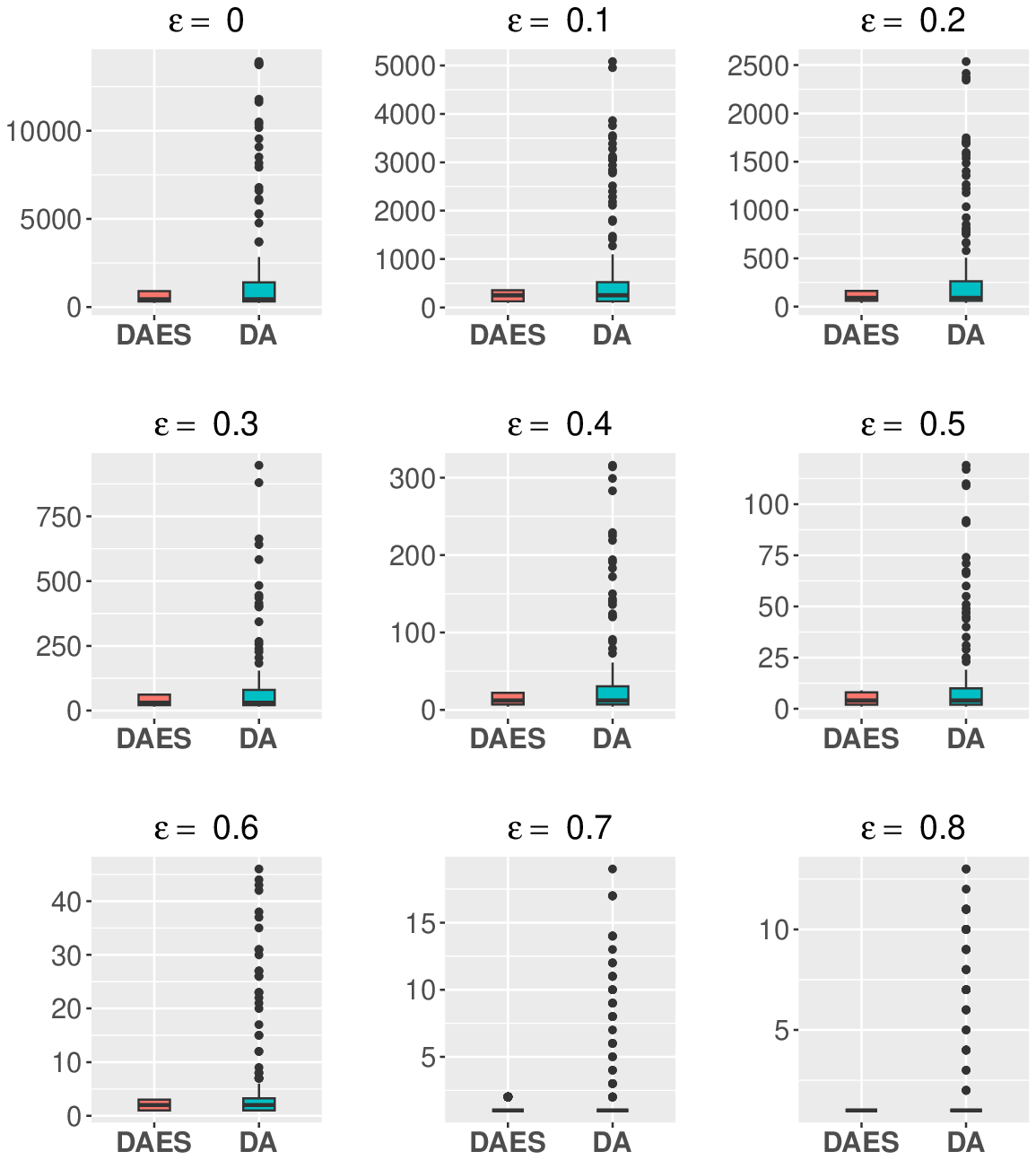}
\end{center}
\caption{Selected  $k_1$ with $\bfX\sim g_1(\bfx)$ and unequally split sub-samples.  }
\label{figure:k:uniform:unequal:x}
\end{figure}

\begin{figure}
\begin{center}
\includegraphics[width=5 in]{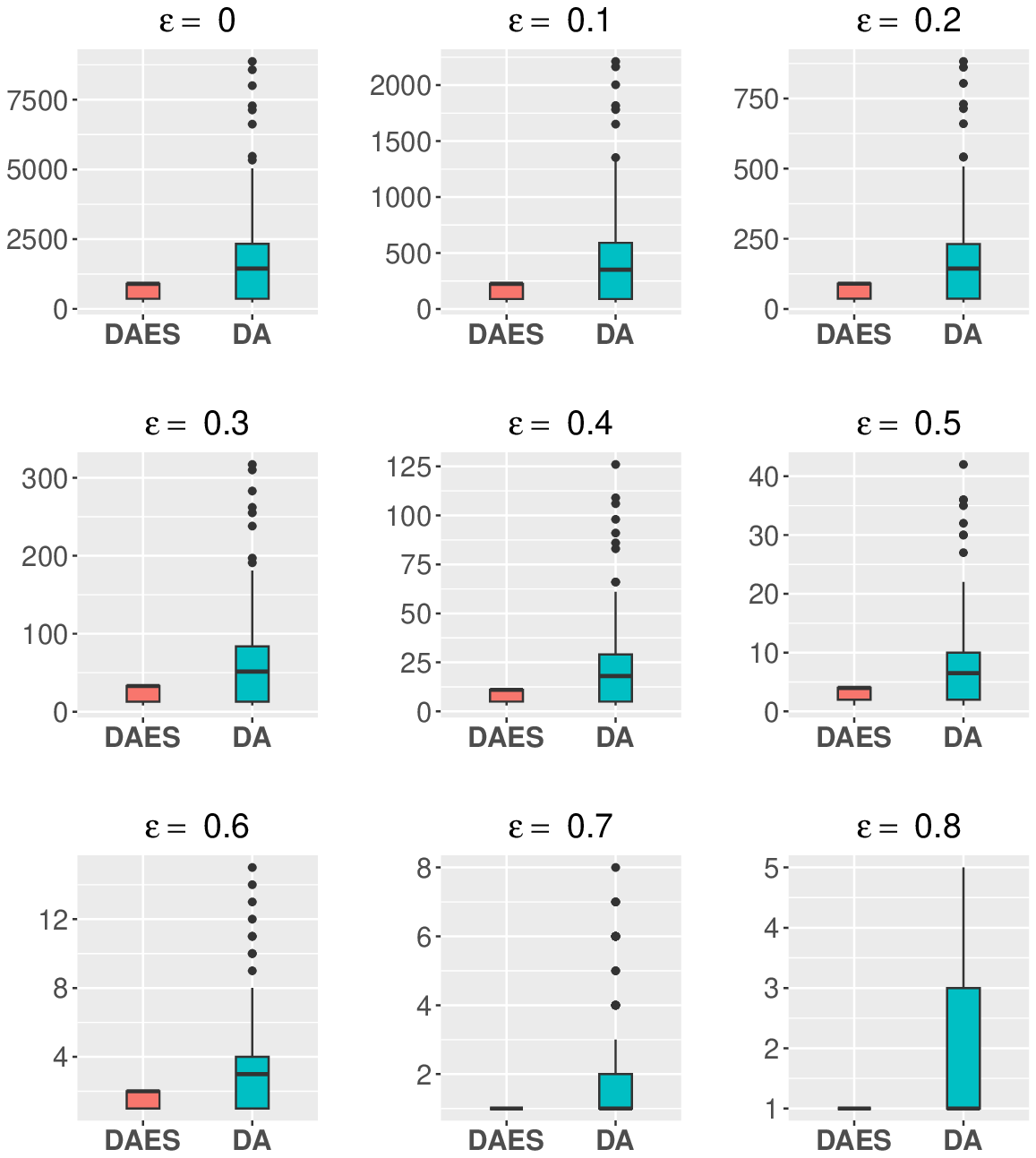}
\end{center}
\caption{Selected $k_1$ with $\bfX\sim g_2(\bfx)$ and equally split sub-samples. }
\label{figure:k:beta:equal:x}
\end{figure}

\begin{figure}
\begin{center}
\includegraphics[width=5 in]{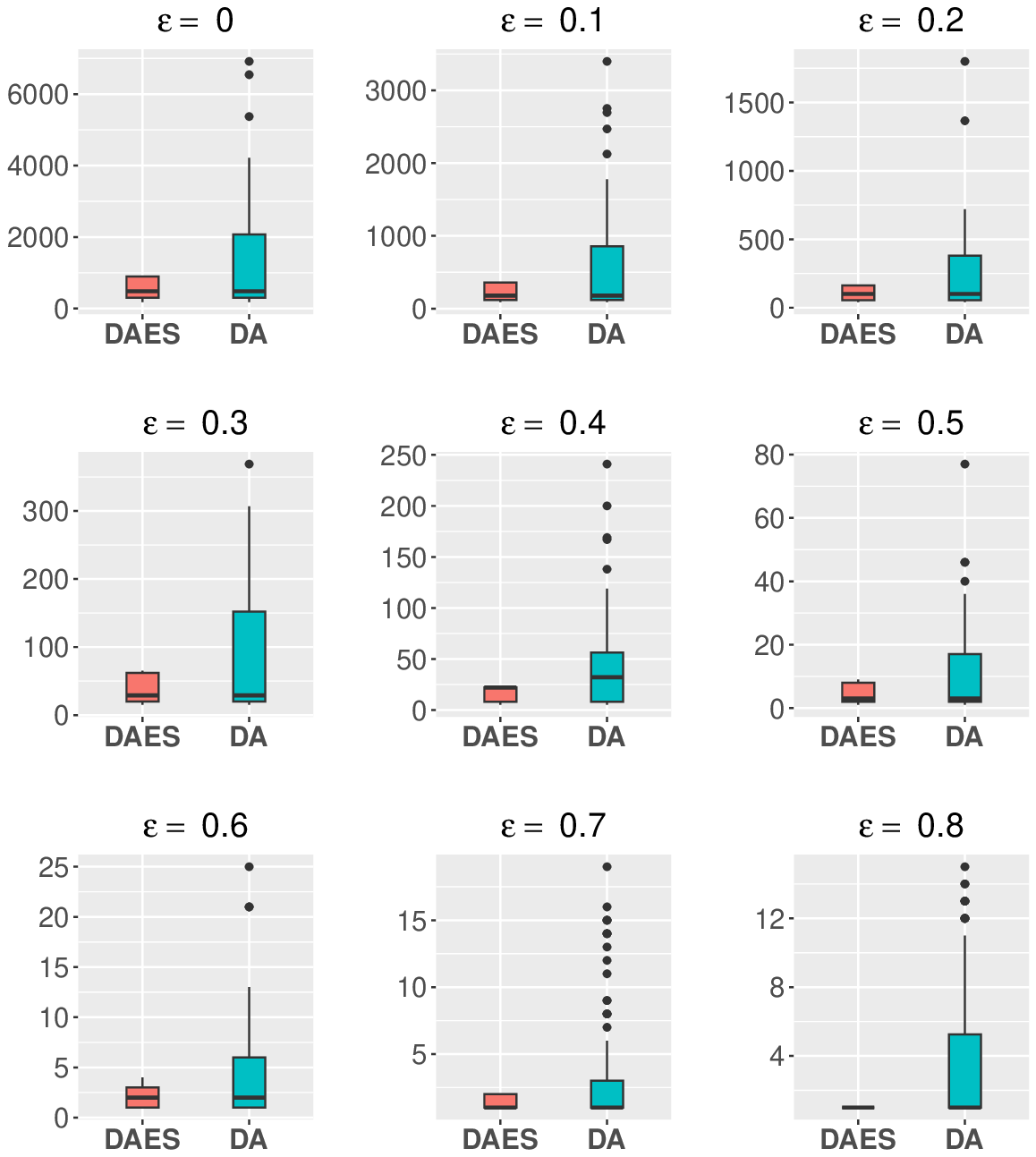}
\end{center}
\caption{Selected  $k_1$ with $\bfX\sim g_2(\bfx)$ and unequally split sub-samples.  }
\label{figure:k:beta:unequal:x}
\end{figure}

For DK in the unequal splitting case, we use  $k_j=\ceil{n_jN^{-\frac{d}{2+d}}}$ for $j=1,\ldots, m$, as suggested by our Theorem \ref{theorem:deterministic:estimator:unequal}. Such a choice reduces to $k=\ceil{n N^{-\frac{d}{2+d}}}$ when $n_1=\cdots=n_m=n$, which is the choice adopted by~\cite{qiao2019rates}. For each simulation run, the above four classifiers are trained using $m$ sub-samples to predict the label of a new feature $\bfx$ randomly generated from the marginal distribution of $\bfX$. To evaluate the classification accuracy, we treat the Bayesian classifier $f^*(\bfx)=\mbI(\eta(\bfx)\geq 1/2)$ as the golden rule and calculate the percentage of times a classifier gives the same prediction as the Bayesian classifier. The average computation times (measured in second and taking $\log$) of DA and DAES with different $\epsilon$ are also recorded. To investigate the role of the early stopping rule, we also compare the numbers of neighbor ($k_1$) chosen by DA and DAES. Summary statistics based on 200 simulation runs are reported in Figures~\ref{figure:risk:time:uniform:x}-\ref{figure:k:beta:unequal:x}.

First, Figures \ref{figure:risk:time:uniform:x} and \ref{figure:risk:time:beta:x} suggest that the proposed DAES classifier has a better overall performance than DK.  In particular, their classification accuracies are practically the same for $\epsilon\geq 0.5$,  while the proposed DAES performs significantly better than  DK  when $\epsilon\leq 0.3$ and $\bfX\sim g_2(\bfx)$, which demonstrates the benefits of searching for an optimal $k$ using a data-driven Algorithm 1. 

A second observation from Figures \ref{figure:risk:time:uniform:x} and \ref{figure:risk:time:beta:x} is that the DAES classifier appears to be consistently inferior to the DA classifier. This highlights the importance of imposing an early stopping bound $\ceil{n_1N^{-\frac{d}{2+d}}}$ when searching for the optimal $k_1$. This can be explained by the fact that searching for $k_1$ from $1$ to $n_1$ may introduce too much uncertainty in the choice of $k_1$ ({as well as other $k_j$'s}), which may, in turn, results in greater variability for the final aggregated NN classifier. This explanation also can be supported by Figures \ref{figure:k:uniform:equal:x}-\ref{figure:k:beta:unequal:x}. For example, Figure \ref{figure:k:uniform:equal:x} shows that the $k_1$ chosen by DA is generally larger than that chosen by DAES. When $\epsilon=0$, DA could choose a $k_1$ larger than $10000$ given $N=60000$, which may increase a lot of uncertainty for classification. 

Third, Figures  Figures \ref{figure:risk:time:uniform:x} and \ref{figure:risk:time:beta:x} also indicate that the proposed DAES classifier performs similarly to the D1 classifier when $\epsilon$ is large, supporting our theoretical findings in Theorems~\ref{theorem:adaptive:estimator}-\ref{theorem:1nn}.  However, the D1 classifier performs much worse than the DAES classifier when $\epsilon$ is small, demonstrating the advantage of the proposed DAES classifier due to its adaptivity in choosing an optimal $k_1$ ({as well as other $k_j$'s}).

Finally, Figures \ref{figure:risk:time:uniform:x} and \ref{figure:risk:time:beta:x} show that for each different marginal distributions of $\bfX$, the DAES classifier outperforms the DA classifier in terms of computation time, both of which are U-shaped functions with respect to $\epsilon$ and attain the minimal when $\epsilon$ is around $0.5$. For small $\epsilon$,  a large proportion of the computation time is spent on choosing $k_1,\ldots, k_m$. However, when $\epsilon$ is large, the main computational cost is to aggregate the sub-samples, resulting in increased run time as $\epsilon$ continues to increase. All numerical studies are conducted via High Performance Computing Center at Texas Tech University. 

\subsection{{A Real Data Analysis}}\label{sec:realdata}
\begin{figure}[htp!]
\centering
\includegraphics[width=2.5 in]{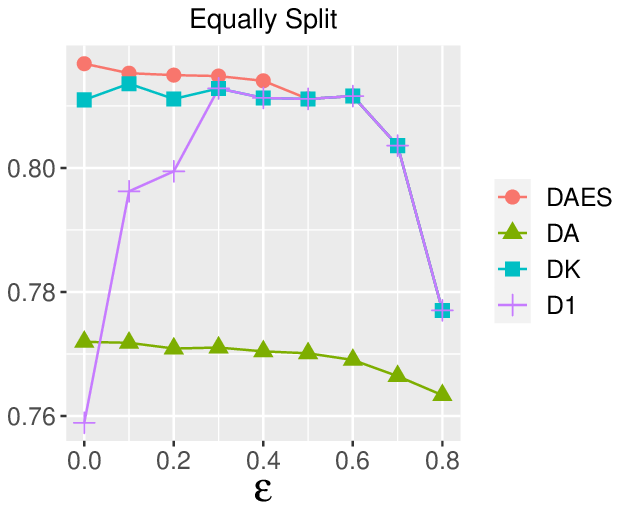}
\includegraphics[width=2.5 in]{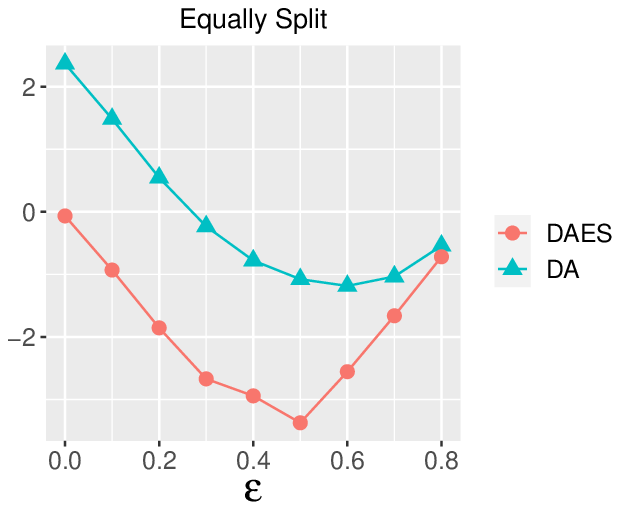}

\includegraphics[width=2.5 in]{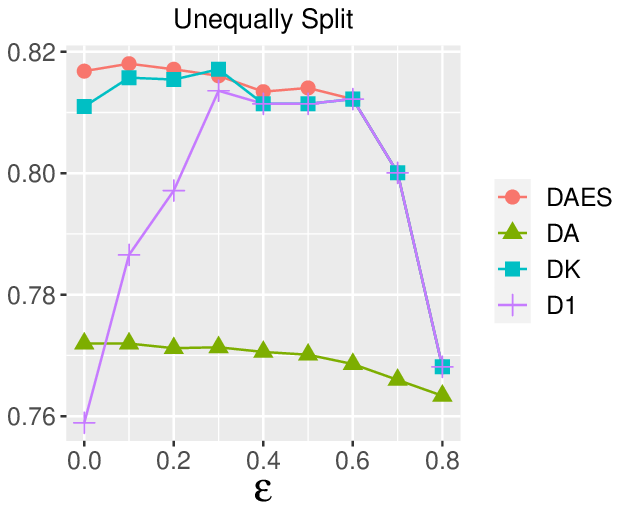}
\includegraphics[width=2.5 in]{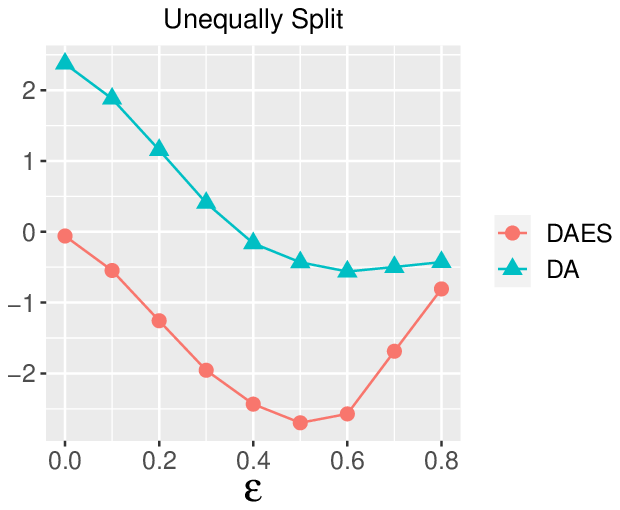}
\caption{Classification accuracy and computation time for adult income dataset. }
\label{figure:adult}
\end{figure}

In this section, we apply the four classifiers in Section \ref{Sec:simulation}  to the adult income dataset from UCI Machine Learning Repository (\citealp{Dua:2019}). The goal is predict  whether a person makes over 50K a year. After removing missing values, we retain $32561$  observations and use age, final weight, education, capital gain, capital loss and weekly working hours  as the feature vector. The whole data is divided into a training dataset with $26049$ observations (about 80\%) and a testing dataset with sample size $6512$ (about 20\%). We use the same settings in Section \ref{Sec:simulation} to evaluate the prediction error of the testing dataset. The results are summarized in Figure \ref{figure:adult}. Overall, our proposed algorithm DAES has the best performance under various choices of $\epsilon$. Moreover, compared with DA, our estimator DAES significantly speeds up the computation when $\epsilon\leq 0.5$.

\section{Conclusion}\label{sec13}
In this work, we study the binary classification problem in the big data setting, and propose a distributed adaptive NN classifier with the tuning parameter being selected by a data-driven criterion. Under mild conditions, we prove the proposed classifier can achieve the minimax optimal rate of excess risk. Numerical results demonstrate its effectiveness and efficiency.

\clearpage
\backmatter

%
%
%
%
\bmhead{Acknowledgments}
The authors would like to express sincere appreciation to Editor Dr. Ricardo Henao and the two anonymous reviewers for their valuable and insightful comments.
%

%
%
%

%
%
%
%

\begin{appendices}

\section{Mathematical Proofs}\label{secA1}
In this Appendix, we provide the mathematical proofs of the theorems and relevant lemmas. 

We denote $\mcX=\{\bfX_1, \bfX_2, \ldots, \bfX_N\}$ as the collection of all covariates. For  $k_j=1,\ldots, n_j$ with $j=1,\ldots, m$ and $0\leq a<1$, we define events
\begin{eqnarray*}
E_j(k_j, a)=\bigg\{\|X_{(k_j)}^j(\bfx)-\bfx\|\leq C_D\bigg(\frac{k_j}{n_j^{1-a}}\bigg)^{\frac{1}{d}}\textrm{ for all } \bfx \in \Omega\bigg\}
\end{eqnarray*}
and
\begin{eqnarray*}
E_P(k_1:k_m, a)=\cap_{j=1}^m E_j(k_j, a).
\end{eqnarray*}
Sometime we may write $E_P(k_1:k_m, a)$ as $E_P$ if there is no confusion in the context. By Lemma \ref{lemma:order:statistics:bound} below, it follows that
\begin{eqnarray}
\pr(E_P(k_1:k_m, a))\geq 1-C_D\sum_{j=1}^m\frac{n_j^{1-a}}{k_j} \exp({-n_j^ak_j/6}).
 \label{eq:probability:Ep}
\end{eqnarray}

\setcounter{section}{0}
\setcounter{equation}{0}
\setcounter{subsection}{0}
\def\theequation{A\arabic{equation}}
\def\thesection{A\arabic{section}}
\def\thesubsection{A\arabic{subsection}}


\subsection{Preliminary Lemmas}
\begin{lemma}\label{lemma:order:statistics:bound}
There exist $C_D>0$ such that for all  $a\in [0, 1)$, $k_j\in \{1,\ldots, n_j\}$ and $j \in \{1,\ldots, m\}$, the following holds with probability at least $1-C_D\frac{n_j^{1-a}}{k_j} e^{-n_j^{a}k_j/6}$:
\begin{eqnarray}
	\|X_{(k_j)}^j(\bfx)-\bfx\|\leq C_D\bigg(\frac{k_j}{n_j^{1-a}}\bigg)^{\frac{1}{d}}\quad \textrm{ for all } \bfx \in \Omega.\nonumber
\end{eqnarray}
Moreover, with probability at least $1-C_D\frac{n_j}{k_j} e^{-k_j/6}$, it also holds that
\begin{eqnarray}
\|X_{(k_j)}^{j}(\bfx)-\bfx\|\geq \frac{1}{C_D}\bigg(\frac{k_j}{n_j}\bigg)^{\frac{1}{d}}\quad \textrm{ for all } \bfx \in \Omega.\nonumber
\end{eqnarray}
\end{lemma}
\begin{proof}
The proofs of the upper bound and lower bound are almost the same. In the following, we prove the upper bound. For simplicity, we will omit the index $j$.

Let $B(\bfx, r)$ be the ball centered at $\bfx$ with radius $r$. By Assumption \ref{A1:strong:density}, therefore we have
\begin{eqnarray*}
\pr(\bfX\in B(\bfx, r))=\int_{B(\bfx, r)\cap \Omega} \frac{dP_\bfX(\bfx)}{d\lambda}(\bfx)d\bfx&\geq&  c_\lambda\lambda(B(\bfx, r)\cap \Omega)\\
&\geq&c_\lambda^{2} \lambda(B(\bfx, r))=c_\lambda^{2} \lambda(B(0, 1))r^d.
\end{eqnarray*} 
For simplicity, we denote  $c=c_\lambda^{2}\lambda(B(0,1))$, so $\pr(\bfX\in B(\bfx, r))\geq cr^d$. Let $r=c_r(k/n^{1-a})^{\frac{1}{d}}$ for some $0\leq a\leq 1$ and $c_r=(2/c)^{\frac{1}{d}}$. Moreover, we define $S(\bfx)=\sum_{i=1}^{n}\mbI(\bfX_i^j\in B(\bfx, r))$ and $W\sim Binomial(n, cr^d)$. Hence, Bernstein's inequality implies that
\begin{eqnarray*}
\pr(S(\bfx)< k)\leq  \pr(W< k)&=&\pr(W-\ev(W)\geq k-\ev(W))\nonumber\\
&=&\pr(W-\ev(W)< k-cnr^d)\nonumber\\
&=&\pr(W-\ev(W)< k-cc_r^dn^{a}k)\nonumber\\
&=&\pr\bigg(W-\ev(W)<k-2n^ak\bigg)\nonumber\\
&\leq &\pr\bigg(W-\ev(W)<-n^ak\bigg)\nonumber\\
&\leq&  \exp\bigg(-\frac{3n^ak}{14}\bigg)\leq \exp(-n^ak/6),
\end{eqnarray*}
where we use the fact that $a\geq 0$.
Let $\mcB\subset \Omega$ be a finite set such that $\Omega \subset \bigcup_{\bfx \in \mcB}B(\bfx, r)$, and we can verify $|\mcB|\leq Cr^{-d}$ for some $C>0$. As a consequence, we have
\begin{eqnarray*}
\pr(\exists \bfx \in \mcB, S(\bfx)< k)\leq Cr^{-d}\exp(-n^ak/6)\leq C c_r^{-d}\frac{n^{1-a}}{k}\exp(-n^ak/6).
\end{eqnarray*}
For any $\bfx\in \Omega$, there is a $\bfx' \in \mcB$ such that $\|\bfx'-\bfx\|\leq 2r$. Under the event $E_2=\{\forall \bfx \in \mcB, S(\bfx)\geq k\}$, there are at least $k$ covariates among $\bfX_1^j,\ldots, \bfX_{n}^j$ in the ball $B(\bfx', r)$, and thus there are at least $k$ covariates among $\bfX_1^j, \ldots, \bfX_{n}^j$ in the ball $B(\bfx, 2r)$. Hence, we have
\begin{eqnarray*}
\pr(\exists \bfx \in \Omega, \|\bfX_{(k)}^j(\bfx)-\bfx\|\leq 2r)\geq \pr(E_2)\geq 1-C c_r^{-d}\frac{n^{1-a}}{k}\exp(-n^ak/6).
\end{eqnarray*}
\end{proof}

\begin{lemma}\label{lemma:bias:bound}
Fixing $k_1=1,\ldots,n_1$  and setting $k_j=\ceil{k_1n_j/n_1}$ with $j=1,\ldots, m$, there exist $c_b, C_b>0$ free of $k_j$ such that 
\begin{eqnarray}
	\begin{cases}
	\ev(\widehat{\eta}_{k_j,j}(\bfx)|\mcX)-\frac{1}{2}\geq c_b\zeta(\bfx) & \textrm{if } f^*(\bfx)=1,\\
	\ev(\widehat{\eta}_{k_j,j}(\bfx)|\mcX)-\frac{1}{2}\leq -c_b\zeta(\bfx) & \textrm{if } f^*(\bfx)=0,\\
	\end{cases}\nonumber
\end{eqnarray}
holds for all $\bfx$ with $\zeta(\bfx)\geq C_b \|\bfX_{(k_j)}^j(\bfx)-\bfx\|^\beta$. Moreover, if $k_1n_j\geq n_1$ for all $j=1,\ldots, m$, then the following statements hold on event $E_P(k_1:k_m, 0)$:
\begin{eqnarray}
	\begin{cases}
	\ev(\widehat{\eta}_{k_1:k_m}(\bfx)|\mcX)-\frac{1}{2}\geq c_b\zeta(\bfx) & \textrm{if } f^*(\bfx)=1,\\
	\ev(\widehat{\eta}_{k_1:k_m}(\bfx)|\mcX)-\frac{1}{2}\leq -c_b\zeta(\bfx) & \textrm{if } f^*(\bfx)=0;\\
	\end{cases}\nonumber
\end{eqnarray}
for all $\bfx$ with $\zeta(\bfx)\geq C_b(k_1/n_1)^{\frac{\beta}{d}}$. In addition, if $k_1=\ldots=k_m=k$ and $n_1=\ldots=n_m=n$, then for any $a\in [0, 1]$, the following statements hold on event $E_P(k_1:k_m, a)$:
\begin{eqnarray}
	\begin{cases}
	\ev(\widehat{\eta}_{k_1:k_m}(\bfx)|\mcX)-\frac{1}{2}\geq c_b\zeta(\bfx) & \textrm{if } f^*(\bfx)=1,\\
	\ev(\widehat{\eta}_{k_1:k_m}(\bfx)|\mcX)-\frac{1}{2}\leq -c_b\zeta(\bfx) & \textrm{if } f^*(\bfx)=0;\\
	\end{cases}\nonumber
\end{eqnarray}
for all $\bfx$ with $\zeta(\bfx)\geq C_b(k/n^{1-a})^{\frac{\beta}{d}}$.
\end{lemma}
\begin{proof}
Since $\ev(Y_{(i)}^j(\bfx)|\mcX)=\eta(\bfX_{(i)}^j(\bfx))$, by Assumption \ref{A1:holder:smooth}, we show that
\begin{eqnarray*}
|\ev(\widehat{\eta}_{k_j, j}(\bfx)|\mcX)-\eta(\bfx)|&=&\bigg|\frac{1}{k_j}\sum_{i=1}^{k_j}[\ev(Y_{(i)}^j(\bfx)|\mcX)-\eta(\bfx)]\bigg|\nonumber\\
&=&\bigg|\frac{1}{k_j}\sum_{i=1}^{k_j}[\eta(\bfX_{(i)}^j(\bfx))-\eta(\bfx)]\bigg|\nonumber\\
&\leq& \frac{C_\beta}{k_j}\sum_{i=1}^{k_j} \|(\bfX_{(i)}^j(\bfx)-\bfx\|^\beta\nonumber\\
&\leq&C_\beta \|(\bfX_{(k_j)}^j(\bfx)-\bfx\|^\beta\nonumber.
\end{eqnarray*}
Therefore, choosing $C_b\geq 2C_\beta$ and if $\zeta(\bfx)\geq C_b \|\bfX_{(k_j)}^j(\bfx)-\bfx\|^\beta=2C_\beta \|\bfX_{(k_j)}^j(\bfx)-\bfx\|^\beta$ and $f^*(\bfx)=1$, then we have
\begin{eqnarray*}
\ev(\widehat{\eta}_{k_j,j }(\bfx)|\mcX)-\frac{1}{2}&\geq& \eta(\bfx)-\frac{1}{2}-|\ev(\widehat{\eta}_{k_j,j}(\bfx)|\mcX)-\eta(\bfx)|\nonumber\\
&\geq& \zeta(\bfx)-C_\beta \|(\bfX_{(k_j)}^j(\bfx)-\bfx\|^\beta\geq \frac{1}{2}\zeta(\bfx).
\end{eqnarray*}
So the statement will hold for $c_b\leq 1/2$ and $C_b\geq 2 C_\beta$. Similarly, we can prove the case when $\zeta(\bfx)\geq C_b \|(\bfX_{(k_j)}^j(\bfx)-\bfx\|^\beta$ and $f^*(\bfx)=0$. Consequently, on event $E_P(k_1:k_m,0)$, we have
\begin{eqnarray*}
|\ev(\widehat{\eta}_{k_1:k_m}(\bfx)|\mcX)-\eta(\bfx)|&=&\bigg|\frac{1}{\sum_{j=1}^mk_j}\sum_{j=1}^m\sum_{i=1}^{k_j}[\ev(Y_{(i)}^j(\bfx)|\mcX)-\eta(\bfx)]\bigg|\nonumber\\
&=&\bigg|\frac{1}{\sum_{j=1}^mk_j}\sum_{j=1}^m\sum_{i=1}^{k_j}[\eta(\bfX_{(i)}^j(\bfx))-\eta(\bfx)]\bigg|\nonumber\\
&\leq& \frac{C_\beta}{\sum_{j=1}^mk_j}\sum_{j=1}^m\sum_{i=1}^{k_j} \|(\bfX_{(i)}^j(\bfx)-\bfx\|^\beta\nonumber\\
&\leq& \frac{C_\beta}{\sum_{j=1}^mk_j}\sum_{j=1}^m k_j\|(\bfX_{(k_j)}^j(\bfx)-\bfx\|^\beta\nonumber\\
&\leq&  \frac{C_\beta C_D^\beta}{\sum_{j=1}^mk_j} \sum_{j=1}^m k_j\bigg(\frac{k_j}{n_j}\bigg)^{\frac{\beta}{d}}\nonumber\\
&\leq&  \frac{C_\beta C_D^\beta}{\sum_{j=1}^mk_j} \sum_{j=1}^m k_j\bigg(\frac{2k_1}{n_1}\bigg)^{\frac{\beta}{d}}\leq  2^{\frac{\beta}{d}}C_\beta C_D^\beta \bigg(\frac{k_1}{n_1}\bigg)^{\frac{\beta}{d}}\nonumber,
\end{eqnarray*}
where the condition $k_j=\ceil{k_1n_j/n_1}\leq 2k_1n_j/n_1$ is used. For $C_b\geq 2^{1+\frac{\beta}{d}}C_\beta C_D^\beta $ and $\bfx$ such that $\zeta(\bfx)\geq C_b(k_1/n_1)^{\frac{\beta}{d}}$, $f^*(\bfx)=1$, it holds that
\begin{eqnarray*}
\ev(\widehat{\eta}_{k_1:k_m}(\bfx)|\mcX)-\frac{1}{2}&\geq& \eta(\bfx)-\frac{1}{2}-|\ev(\widehat{\eta}_{k_1:k_m}(\bfx)|\mcX)-\eta(\bfx)|\nonumber\\
&\geq& \zeta(\bfx)-2^{\frac{\beta}{d}}C_\beta C_D^\beta \bigg(\frac{k_1}{n_1}\bigg)^{\frac{\beta}{d}}\nonumber\\
&\geq& C_b\bigg(\frac{k_1}{n_1}\bigg)^{\frac{\beta}{d}}-\frac{C_b}{2}\bigg(\frac{k_1}{n_1}\bigg)^{\frac{\beta}{d}} \geq \frac{C_b}{2}\bigg(\frac{k_1}{n_1}\bigg)^{\frac{\beta}{d}}.
\end{eqnarray*}
Finally, choosing $c_b\leq C_b/2$, we complete the proof of the second statement. Similarly, we can prove the case when $\zeta(\bfx)\geq C_b(k_1/n_1)^{\frac{\beta}{d}}$ and $f^*(\bfx)=0$.

The proof of the third statement is similar to the second one. Hence, we omit it.
\end{proof}

\begin{lemma}\label{lemma:case:1:exponential:inequality}
Let $c_b$ and $C_b$ be the constants in Lemma \ref{lemma:bias:bound}. Fixing $k_1=1,\ldots,n_1$  and setting $k_j=\ceil{k_1n_j/n_1}$ with $j=1,\ldots, m$,  if $k_1n_j\geq n_1$ for all $j=1,\ldots, m$, then the following holds on event $E_{P}(k_1:k_m, 0)$:  
\begin{eqnarray}
	\pr(\widehat{f}_{k_1:k_m}(\bfx)\neq f^*(\bfx)|\mcX)\leq \exp\bigg(-{2c_b^2 \sum_{j=1}^mk_j\zeta^{2}(\bfx)}\bigg),\nonumber
\end{eqnarray}
for all $\bfx$ with $\zeta(\bfx)\geq C_b(k_1/n_1)^{\frac{\beta}{d}}$.  In addition, if $k_1=\ldots=k_m=k$ and $n_1=\ldots=n_m=n$, then for any $a\in [0, 1]$, the following statements hold on event $E_P(k_1:k_m, a)$:
\begin{eqnarray}
	\pr(\widehat{f}_{k_1:k_m}(\bfx)\neq f^*(\bfx)|\mcX)\leq \exp\bigg(-{2c_b^2 mk\zeta^{2}(\bfx)}\bigg),\nonumber
\end{eqnarray}
for all $\bfx$ with $\zeta(\bfx)\geq C_b(k/n^{1-a})^{\frac{\beta}{d}}$.
\end{lemma}
\begin{proof}
Suppose $f^*(\bfx)=1$ and $\zeta(\bfx)\geq C_b(k_1/n_1)^{\frac{\beta}{d}}$. By Lemma \ref{lemma:bias:bound}, under event $E_{P}(k_1:k_m, 0)$, we have
\begin{eqnarray}
	\ev\bigg(\widehat{\eta}_{k_1:k_m}(\bfx)-\frac{1}{2}\bigg| \mcX\bigg)\geq c_b \zeta(\bfx).\label{eq:lemma:case:1:exponential:inequality:eq:1}
\end{eqnarray}
Furthermore, we observe the that
\begin{eqnarray}
	\widehat{\eta}_{k_1:k_m}(\bfx)=\frac{1}{\sum_{j=1}^m k_j}\sum_{j=1}^{m}\sum_{i=1}^{k_j}Y_{(i)}^{j}(\bfx),\nonumber
\end{eqnarray}
and  $Y_{(1)}^{1}(\bfx),\ldots, Y_{(k_1)}^{1}(\bfx),\ldots, Y_{(1)}^{m}(\bfx),\ldots, Y_{(k_m)}^{m}(\bfx)$ are independent conditional on $\mcX$. Hence, it follows form Hoeffding's inequality and (\ref{eq:lemma:case:1:exponential:inequality:eq:1}) that
\begin{eqnarray}
	&&\pr(\widehat{f}_{k_1:k_m}(\bfx)\neq f^*(\bfx)|\mcX)\nonumber\\
	&=&\pr(\widehat{f}_{k_1:k_m}(\bfx)=0|\mcX)\nonumber\\
	&=&\pr\bigg(\widehat{\eta}_{k_1:k_m}(\bfx)-\frac{1}{2}< 0\bigg|\mcX\bigg)\nonumber\\
	&=&\pr\bigg\{\widehat{\eta}_{k_1:k_m}(\bfx)-\ev(\widehat{\eta}_{k_1:k_m}(\bfx)|\mcX)< -\ev\bigg(\widehat{\eta}_{k_1:k_m}(\bfx)-\frac{1}{2}\bigg|\mcX\bigg)\bigg|\mcX\bigg\}\nonumber\\
	&\leq &\pr\bigg\{\widehat{\eta}_{k_1:k_m}(\bfx)-\ev(\widehat{\eta}_{k_1:k_m}(\bfx)|\mcX)< -c_b\zeta(\bfx)\bigg|\mcX\bigg\}\nonumber\\
	&\leq& \exp\bigg(-{2c_b^2\sum_{j=1}^mk_j \zeta^{2}(\bfx)}\bigg).\nonumber
\end{eqnarray}
Using similar argument, we can prove the case when  $\zeta(\bfx)\geq C_b(k_1/n_1)^{\frac{\beta}{d}}$ and $f^*(\bfx)=0$.
\end{proof}

\subsection{{Proof of Lemma \ref{lemma:counting:vc} } }
\begin{figure}[H]
\centering
\begin{minipage}[b]{0.3\linewidth}
\includegraphics[width=1.5in]{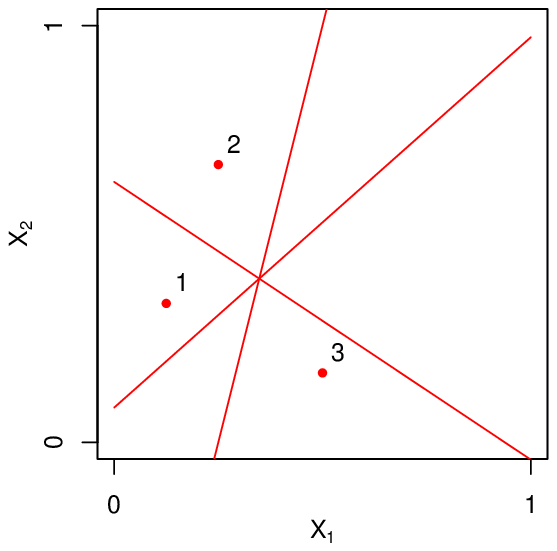}
\caption{The partition determining the possible sets of $A_{3, 1}(\bfx)$ for three points.}
\label{fig:shatter1}
\end{minipage}
\quad
\begin{minipage}[b]{0.3\linewidth}
\includegraphics[width=1.5in]{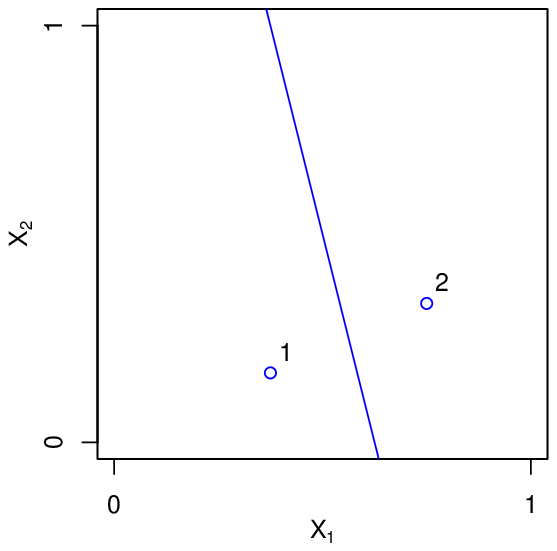}
\caption{The partition determining the possible sets of $A_{2, 2}(\bfx)$ for two points.}
\label{fig:shatter2}
\end{minipage}
\quad
\begin{minipage}[b]{0.3\linewidth}
\includegraphics[width=1.5in]{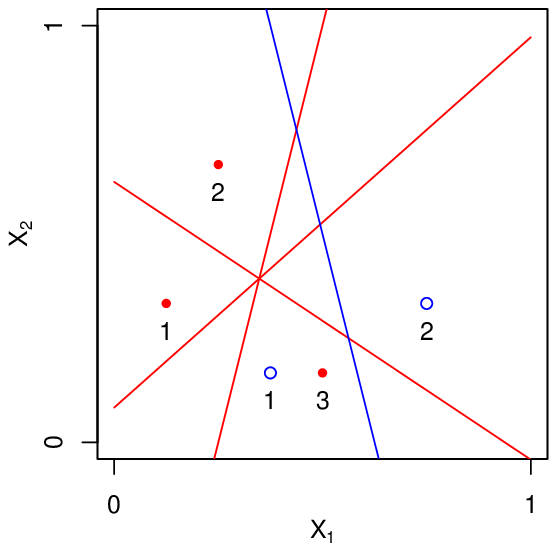}
\caption{The partition determining the possible sets of $A_{3,1}(\bfx)\times A_{2, 2}(\bfx)$.}
\label{fig:shatter3}
\end{minipage}
\end{figure}

Let $\mcH$ be the partition of $[0, 1]^d$ induced by $m{n \choose 2}$ hyperplanes defined as the perpendicular bisectors of each pair of points $(\bfX_s^j, \bfX_p^j)$ for $1\leq s<p\leq n$ and $j=1,\ldots, m$ (see Figure \ref{fig:shatter3} for the case with $m=2, k_1=3, k_2=2$). If $\bfx$ and $\bfx'$ are in the same partition, then $A_{k_j,j}(\bfx)=A_{k_j,j}(\bfx')$ for all $j=1,\ldots, m$ (see Figures \ref{fig:shatter1} and \ref{fig:shatter2}). As a consequence, the cardinality $|\mcB|\leq |\mcH|$. Now consider $\widetilde{\mcH}$ to be the partition of $[0, 1]^d$ induced by ${N \choose 2}$ hyperplanes defined as the perpendicular bisectors of each pair of points $(\bfX, \widetilde{\bfX})$ with $\bfX\neq \widetilde{\bfX}$. Then   $\widetilde{\mcH}$ is a refined partition of $\mcH$, thus $|\mcH|\leq |\widetilde{\mcH}|$.  Now by Lemma 3 in \cite{jiang2019non}, we have $ |\widetilde{\mcH}|\leq dN^d$.

\subsection{{Proof of Theorem \ref{theorem:adaptive:estimator:unequal} } }

In this Section, let us define  set
\begin{eqnarray*}
\Gamma_m=\{(k_1,\ldots, k_m):  k_j=\ceil{k_1n_j/n_1}, k_1=1,\ldots,n_1,  j=1,\ldots, m\}.
\end{eqnarray*}
and quantity
$$\delta=C_\delta[N/\log(N)]^{-\frac{\beta}{2\beta+d}}$$ for some large enough constant $C_\delta$.  For $j=1,\ldots, m$, we denote the following random quantities:
\begin{eqnarray*}
k_j^{\textrm{opt}}(\bfx)=\max\bigg\{k : \|\bfX_{(k)}^{j}(\bfx)-\bfx\|\leq(C_b^{-1}\delta)^{\frac{1}{\beta}}\bigg\}.
\end{eqnarray*}
For simplicity, we may write $k_j^{\textrm{opt}}$ as $k_j^{\textrm{opt}}(\bfx)$ during the proof, if there is no confusion in the context. Define event $E_A$  that (\ref{eq:lemma:uniform:concentration:eta:hat:m:version:eq:1}) holds for all $\bfx\in [0, 1]^d$ and all $(k_1,\ldots, k_m)\in \Gamma_m$. Then  by Lemma \ref{lemma:uniform:concentration:eta:hat:m:version}, we have
\begin{equation}\label{eq:probability:EA}
\pr(E_A)\geq 1-dN^{-1}
\end{equation}

\begin{lemma}\label{lemma:uniform:concentration:eta:hat:m:version}
For any $\epsilon>0$, with probability at least $1-\tau$, the following holds: 
\begin{eqnarray*}
\big|\widehat{\eta}_{k_1:k_m}(\bfx)-\ev(\widehat{\eta}_{k_1:k_m}(\bfx)|\mcX)\big|\leq \sqrt{\frac{(d+1)\log(N)-\log(\tau/d)}{2\sum_{j=1}^mk_j}},
\end{eqnarray*}
$\textrm{ for all } \bfx\in [0, 1]^d \textrm{ and all } (k_1,\ldots, k_m)\in \Gamma_m$. As a consequence, choosing $\tau=dN^{-1}$,  the following holds with probability at least $1-dN^{-1}$:
\begin{eqnarray}
\big|\widehat{\eta}_{k_1:k_m}(\bfx)-\ev(\widehat{\eta}_{k_1:k_m}(\bfx)|\mcX)\big|\leq \sqrt{\frac{(d+2)\log(N)}{2\sum_{j=1}^mk_m}}, \label{eq:lemma:uniform:concentration:eta:hat:m:version:eq:1}
\end{eqnarray}
$\textrm{ for all } \bfx\in [0, 1]^d \textrm{ and all }   (k_1,\ldots, k_m)\in \Gamma_m$.
\end{lemma}
\begin{proof}
Notice that $\widehat{\eta}_{k_1:k_m}(\bfx)=\frac{1}{\sum_{j=1}^m k_j}\sum_{j=1}^m\sum_{i=1}^{k}Y_{(i)}^{j}(\bfx)$, and $Y_{(1)}^{1}(\bfx),\ldots,$ $Y_{(k_1)}^{1}(\bfx)$ $,\ldots,$ $Y_{(1)}^{m}(\bfx),$ $\ldots, Y_{(k_m)}^{m}(\bfx)$ are independent conditional on $\mcX$. Therefore, Hoeffding's inequality implies that
\begin{eqnarray*}
\pr\bigg(\big|\widehat{\eta}_{k_1:k_m}(\bfx)-\ev(\widehat{\eta}_{k_1:k_m}(\bfx)|\mcX)\big|>t\bigg|\mcX\bigg)\leq \exp(-2t^2\sum_{j=1}^m k).
\end{eqnarray*}
Conditioning on $\mcX$, for fixed $(k_1,\ldots, k_m) \in \Gamma_m$, when $\bfx$ is running over $[0, 1]^d$, then by Lemma \ref{lemma:counting:vc}, there are at most $dN^d$ different choices of $Y_{(1)}^{1}(\bfx),\ldots,$ $Y_{(k_1)}^{1}(\bfx)$ $,\ldots,$ $Y_{(1)}^{m}(\bfx),$ $\ldots, Y_{(k_m)}^{m}(\bfx)$. Therefore, it follows that
\begin{eqnarray*}
&&\pr\bigg(\exists  \bfx\in [0, 1]^d  \textrm{ such that } \big|\widehat{\eta}_{k_1:k_m}(\bfx)-\ev(\widehat{\eta}_{k_1:k_m}(\bfx)|\mcX)\big|>t\bigg|\mcX\bigg)\\
&\leq& dN^d\exp\bigg(-2t^2\sum_{j=1}^m k\bigg)\nonumber,
\end{eqnarray*}
which further implies that
\begin{eqnarray*}
&&\pr\bigg(\exists (k_1,\ldots, k_m)\in \Gamma_m,  \bfx\in [0, 1]^d \textrm{ such that } \big|\widehat{\eta}_{k_1:k_m}(\bfx)-\ev(\widehat{\eta}_{k_1:k_m}(\bfx)|\mcX)\big|>t\bigg|\mcX\bigg)\nonumber\\
&\leq& dn_1N^{d}\exp\bigg(-2t^2\sum_{j=1}^mk_j\bigg)\leq d\exp\bigg(-2t^2\sum_{j=1}^mk_j+(d+1)\log(N)\bigg)\nonumber.
\end{eqnarray*}
Plug in $t=\sqrt{\frac{(d+1)\log(N)-\log(\tau/d)}{2\sum_{j=1}^mk_j}}$ into above inequality and take expectation, we complete the proof.
\end{proof}

\begin{lemma}\label{lemma:claim:1}
If $\zeta(\bfx)\geq \delta$ and $k\leq k_j^{\textrm{opt}}(\bfx)$, then it holds that
\begin{eqnarray}
	\begin{cases}
	\ev(\widehat{\eta}_{k,j}(\bfx)|\mcX)-\frac{1}{2}\geq c_b\eta(\bfx) & \textrm{if } f^*(\bfx)=1,\\
	\ev(\widehat{\eta}_{k,j}(\bfx)|\mcX)-\frac{1}{2}\leq -c_b\eta(\bfx) & \textrm{if } f^*(\bfx)=0.\\
	\end{cases}\nonumber
\end{eqnarray}
As a consequence, if $\zeta(\bfx)\geq \delta$  and $k_j\leq k_{j}^{\textrm{opt}}(\bfx)$ for all $j=1,\ldots, m$, then the following holds:
\begin{eqnarray}
	\begin{cases}
	\ev(\widehat{\eta}_{k_1:k_m}(\bfx)|\mcX)-\frac{1}{2}\geq c_b\eta(\bfx) & \textrm{if } f^*(\bfx)=1,\\
	\ev(\widehat{\eta}_{k_1:k_m}(\bfx)|\mcX)-\frac{1}{2}\leq -c_b\eta(\bfx) & \textrm{if } f^*(\bfx)=0.\\
	\end{cases}\nonumber
\end{eqnarray}
\end{lemma}
\begin{proof}
For $k\leq k^{\textrm{opt}}$, we have
\begin{eqnarray}
	\|X_{(k)}^{j}(\bfx)-\bfx\|\leq \|X_{(k_j^{\textrm{opt}})}^{j}(\bfx)-\bfx\|\leq (C_b^{-1}\delta)^{\frac{1}{\beta}},\nonumber  
\end{eqnarray}
which further implies that $\zeta(\bfx)\geq \delta\geq  C_b\|\bfX_{(k)}^{j}(\bfx)-\bfx\|^{\beta}$. Applying Lemma \ref{lemma:bias:bound}, we complete the proof of first statement. The second statement follows from the definition that $\widehat{\eta}_{k_1:k_m}(\bfx)=\sum_{j=1}^mk_j\widehat{\eta}_{k,j}(\bfx)/(\sum_{j=1}^mk_j)$.
\end{proof}

\begin{lemma}\label{lemma:claim:2}
Under event $E_A$, if   $\zeta(\bfx)\geq \delta$ and $\widehat{k}_j(\bfx)\leq k_{j}^{\textrm{opt}}(\bfx)$ for all $j=1,\ldots, m$, then $\widehat{f}_{\widehat{k}:\widehat{k}_m}(\bfx)=f^*(\bfx)$. 
\end{lemma}
\begin{proof}
By definition of $\widehat{k}_1, \ldots, \widehat{k}_m$, we have
\begin{eqnarray}
	|\widehat{\eta}_{\widehat{k}_1:\widehat{k}_m}(\bfx)-1/2|>\sqrt{\frac{(d+2)\log(N)}{2\sum_{j=1}^m\widehat{k}_j}}.\nonumber
\end{eqnarray}
On event $E_A$, it follows that
\begin{eqnarray}
	|\widehat{\eta}_{\widehat{k}_1:\widehat{k}_m}(\bfx)-\ev(\widehat{\eta}_{\widehat{k}_1:\widehat{k}_m}(\bfx)|\mcX)|\leq \sqrt{\frac{(d+2)\log(N)}{2\sum_{j=1}^m\widehat{k}_j}}.\nonumber
\end{eqnarray}
Combining above, we conclude that
\begin{eqnarray}
	|\widehat{\eta}_{\widehat{k}_1:\widehat{k}_m}(\bfx)-1/2|>\bigg|\bigg(\widehat{\eta}_{\widehat{k}_1:\widehat{k}_m}(\bfx)-1/2\bigg)-\bigg(\ev(\widehat{\eta}_{\widehat{k}_1:\widehat{k}_m}(\bfx)|\mcX)-1/2\bigg)\bigg|,\nonumber
\end{eqnarray}
which further implies that
\begin{eqnarray}
	sign\big(\widehat{\eta}_{\widehat{k}_1:\widehat{k}_m}(\bfx)-1/2\big)=sign\big(\ev(\widehat{\eta}_{\widehat{k}_1:\widehat{k}_m}(\bfx)|\mcX)-1/2\big),\nonumber
\end{eqnarray}
for all $\bfx$ with $\zeta(\bfx)\geq \delta$ on event $E_A$. Finally, by Lemma \ref{lemma:claim:1} and above equation, on event $E_A$, if  $\zeta(\bfx)\geq \delta$ and $\widehat{k}_j(\bfx)\leq k_{j}^{\textrm{opt}}(\bfx)$ for all $j=1,\ldots, m$, then
\begin{eqnarray}
	sign\big(\widehat{\eta}_{\widehat{k}_1:\widehat{k}_m}(\bfx)-1/2\big)=\begin{cases}
	1 & \textrm{ if } f^*(\bfx)=1,\\
	-1 & \textrm{ if } f^*(\bfx)=0,\\
	\end{cases}\nonumber
\end{eqnarray}
which completes the proof by noticing that $\widehat{f}_{\widehat{k}_1:\widehat{k}_m}=\mbI(\widehat{\eta}_{\widehat{k}_1:\widehat{k}_m}(\bfx)\geq 1/2)$.
\end{proof}

We are ready to prove Theorem \ref{theorem:adaptive:estimator:unequal}. Let us define the deterministic integers
\begin{eqnarray}
	k_j^*=\ceil*{\frac{1}{2}n_jC_D^{-d} C_b^{-\frac{d}{\beta}}\delta^{\frac{d}{\beta}}}, \quad  k^\dagger=\ceil*{n_jC_D^d C_b^{-\frac{d}{\beta}}\delta^{\frac{d}{\beta}}},\nonumber
\end{eqnarray}
and events
\begin{eqnarray*}
E^*=\bigg\{\|\bfX_{(k_j^*)}^j(\bfx)-\bfx\|\leq C_D\bigg(\frac{k_j^*}{n_j}\bigg)^{\frac{1}{d}}, \textrm{ for all } \bfx\in \Omega \textrm{ and } j=1,\ldots, m\bigg\}
\end{eqnarray*}
and
\begin{eqnarray*}
E^\dagger=\bigg\{\|\bfX_{(k_j^\dagger)}^j(\bfx)-\bfx\|\geq \frac{1}{C_D}\bigg(\frac{k_j^\dagger}{n_j}\bigg)^{\frac{1}{d}}, \textrm{ for all } \bfx\in \Omega \textrm{ and } j=1,\ldots, m\bigg\}
\end{eqnarray*}
Since $1-\epsilon>\frac{d}{2\beta+d}$, so $n_j\delta^{\frac{d}{\beta}}=C_\delta^{\frac{d}{\beta}} n_jN^{-\frac{d}{2\beta+d}}[\log(N)]^{\frac{d}{2\beta+d}}\geq C_\delta^{\frac{d}{\beta}} N^{1-\epsilon-\frac{d}{2\beta+d}}[\log(N)]^{\frac{d}{2\beta+d}}$ is diverging. Without loss of generality, we may assume $k_j^*\geq 1$ and $k_j^\dagger \geq 1$. On event $E^*$, it follows from the definition of $k_j^{\textrm{opt}}$ that
\begin{eqnarray}
\|\bfX_{(k_j^*)}^j(\bfx)-\bfx\|\leq C_D\bigg(\frac{k_j^*}{n_j}\bigg)^{\frac{1}{d}}&\leq& C_D\bigg(\frac{n_jC_D^{-d} C_b^{-\frac{d}{\beta}}\delta^{\frac{d}{\beta}}}{n_j}\bigg)^{\frac{1}{d}}\nonumber\\
&\leq& (C_b^{-1}\delta)^{\frac{1}{\beta}}<\|X_{(k_j^{\textrm{opt}}+1)}^j(\bfx)-\bfx\|.\nonumber
\end{eqnarray}
which further implies that
\begin{eqnarray}
	k_j^{\textrm{opt}}(\bfx)\geq k_j^*\geq \frac{1}{2}n_j C_D^{-d} C_b^{-\frac{d}{\beta}}\delta^{\frac{d}{\beta}}\quad \textrm{ for all }  \bfx\in \Omega \textrm{ and } j=1,\ldots, m.\label{eq:lemma:claim:3:eq:1:unequal}
\end{eqnarray}	
Moreover, on event $E^\dagger$, it also holds that
\begin{eqnarray}
\|\bfX_{(k_j^\dagger)}^j-\bfx\|\geq\frac{1}{C_D}\bigg(\frac{k_j^\dagger}{n_j}\bigg)^{\frac{1}{d}}&\geq& \frac{1}{C_D}\bigg(\frac{n_jC_D^dC_b^{-\frac{d}{\beta}}\delta^{\frac{d}{\beta}}}{n_j}\bigg)^{\frac{1}{d}}\nonumber\\
&=& (C_b^{-1}\delta)^{\frac{1}{\beta}}\geq \|X_{(k_j^{\textrm{opt}})}^j-\bfx\|,\nonumber
\end{eqnarray}
where the last equation follows from the definition of $k_j^{\textrm{opt}}$.  Above inequality  implies that
\begin{eqnarray}
	 k_j^{\textrm{opt}}(\bfx)\leq k_j^\dagger \leq n_j C_D^d C_b^{-\frac{d}{\beta}}\delta^{\frac{d}{\beta}}\label{eq:lemma:claim:3:eq:2:unequal}
\end{eqnarray}	
for all $\bfx \in\Omega \textrm{ and } j=1,\ldots, m.$
Combining (\ref{eq:lemma:claim:3:eq:1:unequal}) and (\ref{eq:lemma:claim:3:eq:2:unequal}), we conclude that on event $E^*\cap E^\dagger$, the following holds:
\begin{eqnarray}
\frac{1}{2} n_j C_D^{-d} C_b^{-\frac{d}{\beta}}\delta^{\frac{d}{\beta}}\leq k_j^{\textrm{opt}}(\bfx)\leq n_j C_D^d C_b^{-\frac{d}{\beta}}\delta^{\frac{d}{\beta}} \label{eq:lemma:claim:3:eq:3:unequal}
\end{eqnarray}
for all $\bfx \in\Omega \textrm{ and } j=1,\ldots, m.$
If we define $k^{\min}_1=\floor{\min\{k_1^{\textrm{opt}}, k_2^{\textrm{opt}}n_1/n_2,\ldots, k_m^{\textrm{opt}}n_1/n_m\}}$ and $k_j^{\min}=\ceil{k_1^{\min}n_j/n_1}$, then we can show that
\begin{eqnarray}
k_j^{\min}= \ceil*{\frac{k_1^{\textrm{opt}}n_j}{n_1}}\leq   \ceil*{\frac{k_j^{\textrm{opt}}n_1n_j}{n_jn_1}}=\ceil{k_j^{\textrm{opt}}}=k_j^{\textrm{opt}}.\label{eq:lemma:claim:3:eq:3.5:unequal}
\end{eqnarray}

 Hence, by (\ref{eq:lemma:claim:3:eq:3:unequal}), it holds on event $E^*\cap E^\dagger$ that
\begin{eqnarray}
\frac{1}{4} n_j C_D^{-d} C_b^{-\frac{d}{\beta}}\delta^{\frac{d}{\beta}}\leq k_j^{\min}(\bfx)\leq 2n_j C_D^d C_b^{-\frac{d}{\beta}}\delta^{\frac{d}{\beta}}\label{eq:lemma:claim:3:eq:4:unequal}
\end{eqnarray}
for all $\bfx \in\Omega \textrm{ and } j=1,\ldots, m.$
By definition, it follows that $k_j^{\min}(\bfx)\leq k_j^{\textrm{opt}}$. So under event $E^*(a)\cap E^\dagger$ , for all $\bfx$ with $\zeta(\bfx)\geq \delta$ and $f^*(\bfx)=1$, Lemma \ref{lemma:claim:1} and (\ref{eq:lemma:claim:3:eq:4:unequal}) together imply that
\begin{eqnarray}
	&&\sqrt{\sum_{j=1}^mk_j^{\min}}\bigg(\ev(\widehat{\eta}_{k_1^{\min}:k_1^{\min}}(\bfx)|\mcX)-\frac{1}{2}\bigg)\nonumber\\
	&\geq& \nonumber \sqrt{\frac{1}{4} \sum_{j=1}^m n_j C_D^{-d} C_b^{-\frac{d}{\beta}}\delta^{\frac{d}{\beta}}} c_b\zeta(\bfx)\\
	&\geq& \sqrt{\frac{1}{4}c_b^2C_D^{-d} C_b^{-\frac{d}{\beta}}}\sqrt{N\delta^{\frac{d}{\beta}+2}} \nonumber\\
	&=& \sqrt{\frac{1}{4}c_b^2C_D^{-d} C_b^{-\frac{d}{\beta}}C_\delta^{\frac{2\beta+d}{\beta}}}\sqrt{N\bigg({N}/{\log(N)}\bigg)^{-1}} \nonumber\\
	&\geq&3\sqrt{\frac{(d+2)\log(N)}{2}},\nonumber
\end{eqnarray}
where the last inequality follows if we choose $C_\delta$ large. By above inequality, on event $E_A\cap E^*\cap E^\dagger$, for all $\bfx$ with $\zeta(\bfx)\geq \delta$ and $f^*(\bfx)=1$, it follows that
 \begin{eqnarray}
	&&\sqrt{\sum_{j=1}^mk_j^{\min}}\bigg(\ev(\widehat{\eta}_{k_1^{\min}:k_1^{\min}}(\bfx)|\mcX)-\frac{1}{2}\bigg)\nonumber\\
	&\geq&\sqrt{\sum_{j=1}^mk_j^{\min}}\bigg(\ev(\widehat{\eta}_{k_1^{\min}:k_m^{\min}}(\bfx)|\mcX)-\frac{1}{2}\bigg)\nonumber\\
	&&-\sqrt{\sum_{j=1}^mk_j^{\min}}\bigg|\widehat{\eta}_{  k_1^{\min}:k_m^{\min} }(\bfx)-\ev(\widehat{\eta}_{k_1^{\min}:k_m^{\min}}(\bfx)|\mcX)\bigg|\nonumber\\
	&\geq&3\sqrt{\frac{(d+2)\log(N)}{2}}-\sqrt{\frac{(d+2)\log(N)}{2}}\nonumber\\
	&>& \sqrt{\frac{(d+2)\log(N)}{2}}.\nonumber
\end{eqnarray}
Similarly, on event $E_A\cap E^*\cap E^\dagger$, for all $\bfx$ with $\zeta(\bfx)\geq \delta$ and $f^*(\bfx)=0$, we can show that
\begin{align}
	\sqrt{\sum_{j=1}^mk_j^{\min}}\bigg(\ev(\widehat{\eta}_{k_1^{\min}:k_1^{\min}}(\bfx)|\mcX)-\frac{1}{2}\bigg)<- \sqrt{\frac{(d+2)\log(N)}{2}}.\nonumber
\end{align}
Therefore, we prove that on event $E_A\cap E^*\cap E^\dagger$, the following holds:
\begin{eqnarray}
		\sqrt{\sum_{j=1}^mk_j^{\min}}\bigg|\widehat{\eta}_{k_1^{\min}:k_m^{\min}}(\bfx)-\frac{1}{2}\bigg|> \sqrt{\frac{(d+2)\log(N)}{2}}\label{eq:lemma:claim:3:eq:4.5:unequal}
\end{eqnarray}
for all $\bfx \textrm{ with } \zeta(\bfx)\geq \delta.$
Now by (\ref{eq:lemma:claim:3:eq:4:unequal}), on event $E_A\cap E^*\cap E^\dagger$, we have
\begin{eqnarray}
k_1^{\min}(\bfx)&\leq&  n_1 C_D^{d} C_b^{-\frac{d}{\beta}}\delta^{\frac{d}{\beta}}\nonumber\\
&=&C_D^{d} C_b^{-\frac{d}{\beta}}C_\delta^{\frac{d}{\beta}} n_1 \bigg({N}/{\log(N)}\bigg)^{-\frac{d}{2\beta+d}}\nonumber\\
&=&C_D^{d} C_b^{-\frac{d}{\beta}}C_\delta^{\frac{d}{\beta}} n_1 N^{-\frac{d}{2\beta+d}}[\log(N)]^{\frac{d}{2\beta+d}}\nonumber\\
&\leq& n_1 N^{-\frac{d}{2+d}}\log(N),\label{eq:lemma:claim:3:eq:5:unequal}
\end{eqnarray}
for all $\bfx$ with $\zeta(\bfx)\geq \delta$ and sufficiently large $N$.  In the following, we will calculate the probability of $E_A\cap E^*\cap E^\dagger$. Since $\alpha \beta\leq d$, it follows that
\begin{eqnarray*}
 \frac{\beta(1+\alpha)}{2\beta+d}= \frac{\beta+\alpha\beta}{2\beta+d}\leq \frac{\beta+d}{2\beta+d}< 1.
\end{eqnarray*}
Using the inequality above and (\ref{eq:probability:EA}), it follows that
\begin{eqnarray}
	\pr(E_A)\geq 1-dN^{-1}\geq 1-dC_\delta^{1+\alpha}N^{-\frac{\beta(1+\alpha)}{2\beta+d}}[\log(N)]^{\frac{\beta(1+\alpha)}{2\beta+d}}=1-d\delta^{1+\alpha},\nonumber	
\end{eqnarray}
for sufficiently large $N$.  Moreover, by Lemma \ref{lemma:order:statistics:bound}, (\ref{eq:lemma:claim:3:eq:1:unequal}) and (\ref{eq:lemma:claim:3:eq:2:unequal}), it follows that 
\begin{eqnarray}
	\pr(E^*)&\geq& 1-C_D\sum_{j=1}^m\frac{n_j}{k_j^*}\exp(-k_j^*/6)\nonumber\\
	&\geq&1-C_DN\exp\bigg(-\frac{n_j\delta^{\frac{d}{\beta}}}{12C_D^{d} C_b^{\frac{d}{\beta}}}\bigg)\nonumber\\
		&\geq &1-C_DN\exp\bigg\{-\frac{1}{12C_D^{d} C_b^{\frac{d}{\beta}}}N^{1-\epsilon}\bigg({N}/{\log(N)}\bigg)^{-\frac{d}{2\beta+d}}\bigg\}\nonumber\\
		&=&1-C_DN\exp\bigg\{-\frac{1}{12C_D^{d} C_b^{\frac{d}{\beta}}}N^{1-\epsilon-\frac{d}{2\beta+d}}[\log(N)]^{\frac{d}{2\beta+d}}\bigg\}\nonumber\\
		&\gtrsim& 1-\delta^{1+\alpha},\nonumber
\end{eqnarray}
where we use the fact that $1-\epsilon>d/(2\beta+d)$. Similarly, we can show that $\pr(E^\dagger)\gtrsim 1-\delta^{1+\alpha}$. Combining above, we show that 
\begin{eqnarray}
\pr(E_A\cap E^* \cap E^\dagger)\geq 1-\pr(E_A)-\pr(E^*)-\pr(E^\dagger)\gtrsim 1-\delta^{1+\alpha}.  \label{eq:lemma:claim:3:eq:5.5:unequal}
\end{eqnarray}
 By (\ref{eq:lemma:claim:3:eq:3.5:unequal}), (\ref{eq:lemma:claim:3:eq:4.5:unequal}) and the definition of $\widehat{k}_1(\bfx),\ldots, \widehat{k}_m(\bfx)$, the following holds on event  $E_A\cap E^*\cap E^\dagger$:
\begin{eqnarray}
 \widehat{k}_j(\bfx)\leq k_j^{\min}(\bfx)\leq k_j^{opt}(\bfx)\label{eq:lemma:claim:3:eq:6:unequal}
\end{eqnarray}
for all $\bfx \textrm{ with } \zeta(\bfx)\geq \delta \textrm{ and } j=1,\ldots, m.$
By (\ref{eq:lemma:claim:3:eq:5:unequal}), (\ref{eq:lemma:claim:3:eq:6:unequal}) and  Lemma \ref{lemma:claim:2}, we can see, it holds on event $E_A\cap E^*\cap E^\dagger$ that: 
\begin{eqnarray}
	\widehat{f}_{\widehat{k}_1:\widehat{k}_m}(\bfx)=f^*(\bfx)\quad \textrm{ and } \quad   \widehat{k}_1(\bfx)\leq n_1N^{-\frac{d}{2+d}}\log(N) \label{eq:lemma:claim:3:eq:7:unequal}
\end{eqnarray}
for all $\bfx \textrm{ with } \zeta(\bfx)\geq \delta.$
By (\ref{eq:lemma:claim:3:eq:7:unequal}), on event $E_A\cap E^*\cap E^\dagger$, $\widehat{f}_{\widehat{k}_1:\widehat{k}_m}(\bfX)\neq f^*(\bfX),$ implies $\zeta(x)<\delta$. As a consequence of (\ref{eq:lemma:claim:3:eq:5.5:unequal}) and Assumption \ref{A1:marginal:assumption}, we have
\begin{eqnarray}
	&&\mcR(\widehat{f}_{\widehat{k}_1:\widehat{k}_m})\nonumber\\
	&=&\ev(\zeta(\bfX)\mbI(\widehat{f}_{\widehat{k}_1:\widehat{k}_m}(\bfX)\neq f^*(\bfX)))\nonumber\\
	&\leq &\ev(\zeta(\bfX)\mbI(\widehat{f}_{\widehat{k}_1:\widehat{k}_m}(\bfX)\neq f^*(\bfX), E_A\cap E^*\cap E^\dagger))+\pr((E_A\cap E^*\cap E^\dagger)^c)\nonumber\\
	&=&\ev(\zeta(\bfX)\mbI(\widehat{f}_{\widehat{k}_1:\widehat{k}_m}(\bfX)\neq f^*(\bfX), E_A\cap E^*\cap E^\dagger, \zeta(\bfX)<\delta))+\pr((E_A\cap E^*\cap E^\dagger)^c)\nonumber\\
	&\lesssim&\delta \pr(\zeta(\bfX)<\delta)+\delta^{1+\alpha}\nonumber\\
	&\lesssim&\delta^{1+\alpha}.\label{eq:lemma:claim:3:eq:8:unequal}
\end{eqnarray}

\newpage

\subsection{Proof of Theorem \ref{theorem:deterministic:estimator:unequal}}
By the conditions given, we have $k_1\geq n_1N^{-\frac{d}{2\beta+d}}\geq N^{1-\epsilon-\frac{d}{2\beta+d}}=N^{\frac{2\beta}{2\beta+d}-\epsilon}\to \infty$ and 
 $k_1n_j/n_1\geq n_jN^{-\frac{d}{2\beta+d}}\geq N^{1-\epsilon-\frac{d}{2\beta+d}}=N^{\frac{2\beta}{2\beta+d}-\epsilon}\to \infty$. Without loss of generality, we assume $k_1n_j/n_1\geq 1$ for all $j=1,\ldots, m$. Let $C_\upsilon$ large enough constant such that $\upsilon=C_\upsilon(k_1/n_1)^{\frac{\beta}{d}} \geq C_b(k_1/n_1)^{\frac{\beta}{d}}$. We further define  sets $A_0=\{\bfx : |\eta(\bfx)-1/2|\leq \upsilon\}$ and $A_j=\{\bfx : 2^{j-1}\upsilon<|\eta(\bfx)-1/2|\leq 2^j\upsilon\}$ for $j\geq 1$. For simplicity, let us  write $E_P(k_1:k_m,0)$  as $E_P$ and $\sum_{j=1}^mk_j/m$ as $\bar{k}$.

If $j=0$, then Assumption \ref{A1:marginal:assumption} shows that
\begin{eqnarray}
	\ev\bigg(|2\eta(\bfX)-1|\mbI(\widehat{f}_{k_1:k_m}(\bfX)\neq f^*(\bfX))\mbI(\bfX\in A_0, E_P)\bigg)&\leq& 2\upsilon \pr(\bfX\in A_0)\nonumber\\
	&\leq& 2C_\alpha \upsilon^{1+\alpha}.\nonumber
\end{eqnarray}

If $j\geq 1$, Assumption \ref{A1:marginal:assumption} and Lemma \ref{lemma:case:1:exponential:inequality} imply that
\begin{eqnarray}
	&&\ev\bigg(|2\eta(\bfX)-1|\mbI(\widehat{f}_{k_1:k_m}(\bfX)\neq f^*(\bfX))\mbI(\bfX\in A_j, E_P)\bigg)\nonumber\\
	&\leq&2^{j+1}\upsilon \ev\bigg\{\mbI(\bfX\in A_j,  E_P)\pr(\widehat{f}_{k_1:k_m}(\bfX)\neq f^*(\bfX) | \bfX, \mcX)\bigg\}\nonumber\\
	&\leq& 2^{j+1}\upsilon \ev\bigg\{\mbI(\bfX\in A_j, E_P)\exp\bigg(-{2c_b^2 m\bar{k}\zeta^{2}(\bfx)}\bigg)\bigg\}\nonumber\\
		&\leq& 2^{j+1}\upsilon \ev\bigg\{\mbI(\bfX\in A_j, E_P)\exp\bigg(-{2c_b^2m\bar{k} 4^{j-1}\upsilon^{2}}\bigg)\bigg\}.\nonumber
\end{eqnarray}
By conditions given, it follows that
\begin{eqnarray}
m\bar{k}=\sum_{j=1}^mk_j\geq \sum_{j=1}^m \frac{k_1n_j}{n_1}\geq \sum_{j=1}^m n_j N^{-\frac{d}{2\beta+d} }=N^{\frac{2\beta}{2\beta+d} }\nonumber.
\end{eqnarray}
and $\upsilon=C_\upsilon (k_1/n_1)^{\frac{\beta}{d}}\geq C_\upsilon N^{-\frac{\beta}{2\beta+d}}$. Therefore, it follows that
\begin{eqnarray}
&&\ev\bigg(|2\eta(\bfX)-1|\mbI(\widehat{f}_{k_1:k_m}(\bfX)\neq f^*(\bfX))\mbI(\bfX\in A_j, E_P)\bigg)\nonumber\\
&\leq& 2^{j+1}\upsilon \ev\bigg\{\mbI(\bfX\in A_j, E_P)\exp\bigg(-{\frac{1}{2}c_b^2 C_\upsilon^2 4^{j} N^{\frac{2\beta}{2\beta+d} } N^{-\frac{2\beta }{2\beta+d}} }\bigg)\bigg\}\nonumber\\
&\leq & 2^{j+1}\upsilon \ev\bigg\{\mbI(\bfX\in A_j, E_P)\exp\bigg(-{\frac{1}{2}c_b^2 C_\upsilon^2  4^{j} }\bigg)\bigg\}\nonumber\\
&\leq & 2^{j+1}\upsilon \pr(\bfX\in A_j)\exp\bigg(-{\frac{1}{2}c_b^2 C_\upsilon^2  4^{j} }\bigg)\nonumber\\
&\leq & 2^{j+1}\upsilon C_\alpha (2^j \upsilon)^\alpha\exp\bigg(-{\frac{1}{2}c_b^2 C_\upsilon^2  4^{j} }\bigg)\nonumber\\
&\leq & 2^{j+1}\upsilon C_\alpha (2^j \upsilon)^\alpha\exp\bigg(-{\frac{1}{2}c_b^2 C_\upsilon^2 4^{j} }\bigg)\nonumber\\
&\leq& 2C_\alpha \upsilon^{1+\alpha} 2^{j(1+\alpha)}\exp\bigg(-{\frac{1}{2}c_b^2 C_\upsilon^2  4^{j} }\bigg).\nonumber
\end{eqnarray}
Taking summation ans using the fact that $\sum_{j=1}^\infty b^j \exp(-c4^j)<\infty$ for all $b,c>0$, we have
\begin{eqnarray}
&&\sum_{j=1}^\infty\ev\bigg(|2\eta(\bfX)-1|\mbI(\widehat{f}_{k_1:k_m}(\bfX)\neq f^*(\bfX))\mbI(\bfX\in A_j, E_P)\bigg)\nonumber\\
&\leq&  2C_\alpha \upsilon^{1+\alpha} \sum_{j=1}^\infty 2^{j(1+\alpha)}\exp\bigg(-{\frac{1}{2}c_b^2 C_\upsilon^2 4^{j} }\bigg)\lesssim \upsilon^{1+\alpha}.\nonumber
\end{eqnarray}
Combining the bounds above, it follows that
\begin{eqnarray}
	\mcR(\widehat{f}_{k_1:k_m})&=&\ev\big(|2\eta(\bfX)-1|\mbI(\widehat{f}_{k_1:k_m}(\bfX)\neq f^*(\bfX))\big)\nonumber\\
	&\leq&\ev\big(|2\eta(\bfX)-1|\mbI(\widehat{f}_{k_1:k_m}(\bfX)\neq f^*(\bfX), E_P^c)\big)\nonumber\\
	&&+\ev\big(|2\eta(\bfX)-1|\mbI(\widehat{f}_{k_1:k_m}(\bfX)\neq f^*(\bfX), \bfX\in A_0,E_P)\big)\nonumber\\
	&&+\sum_{j=1}^\infty\ev\big(|2\eta(\bfX)-1|\mbI(\widehat{f}_{k_1:k_m}(\bfX)\neq f^*(\bfX), \bfX\in A_j,E_P)\big)\nonumber\\
&\lesssim& \pr(E_P^c)+\upsilon^{1+\alpha}\nonumber\\
&\lesssim&\pr(E_P^c)+N^{-\frac{ \beta(1+\alpha)}{2\beta+d}}.\nonumber
\end{eqnarray}
By (\ref{eq:probability:Ep}) and the fact that $k_j\geq k_1n_j/n_1\geq n_jN^{-\frac{d}{2\beta+d}}\geq N^{1-\epsilon-\frac{d}{2\beta+d}}=N^{\frac{2\beta}{2\beta+d}-\epsilon}$, it yields that
\begin{eqnarray*}
\pr(E_P^c)\leq C_D\sum_{j=1}^m\frac{n_j}{k_j} \exp({-k_j/6})\lesssim N^{-\frac{\gamma \beta(1+\alpha)}{d}}.
\end{eqnarray*}
Combining the above two inequalities, we complete the proof.

\subsection{Proof of Theorem \ref{theorem:adaptive:estimator}}
In this section, we consider the equal-size sub-samples. For simplicity, let us assume $n_1=\ldots=n_m=n=N^{1-\epsilon}$ and $m=N/n=N^\epsilon$. Notice that $k_1=\ldots=k_m$ in this setting, so we rewrite $\widetilde{\eta}_{k_1:k_m}$ as $\widehat{\eta}_{k}$ and $\widetilde{f}_{k_1:k_m}$ as $\widehat{f}_{k}$ if $k_1=\ldots=k_m=k$. Moreover, we also define $\widehat{k}=\widehat{k}_1=\ldots=\widehat{k}_m$ to be the data-driven quantities in Algorithm \ref{alg:distributed}. Since Theorem \ref{theorem:adaptive:estimator:unequal} studies the case with $\epsilon<\frac{2\beta}{2\beta+d}$, we will focus on the case with $\epsilon\geq \frac{2\beta}{2\beta+d}$. Abusing notation, let us define  quantities
\begin{eqnarray}
v^*:={v}^*(a,\epsilon)=\frac{(1-\epsilon)(1-a)\beta}{d}\quad \textrm{ if } \epsilon \geq \frac{2\beta}{2\beta+d}, 0<a<1\nonumber
\end{eqnarray}
and $${\delta}=C_{{\delta}}[N/\log(N)]^{-{v}^*}$$ for some large enough constant $C_{{\delta}}$.  For $j=1,\ldots, m$, we denote the following random quantities:
\begin{eqnarray*}
{k}_j^{\textrm{opt}}(\bfx)=\max\bigg\{k : \|\bfX_{(k)}^{j}(\bfx)-\bfx\|\leq(C_b^{-1}\delta)^{\frac{1}{\beta}}\bigg\}.
\end{eqnarray*}
For simplicity, we may write  $v^*$ as $v^*(a,\epsilon)$ and $k_j^{\textrm{opt}}$ as $k_j^{\textrm{opt}}(\bfx)$ during the proof, if there is no confusion in the context.  Clearly, Lemmas \ref{lemma:claim:1} and \ref{lemma:claim:2} are still valid under the new $\delta$ and ${k}_j^{\textrm{opt}}(\bfx)$.

\begin{lemma}\label{lemma:claim:3}
Suppose $\frac{2\beta}{2\beta+d}\leq \epsilon\leq \frac{2}{2+d}$, then there exists a constant $c>0$ such that the following holds
\begin{eqnarray*}
\mcR(\widehat{f}_{\widehat{k}})\lesssim\bigg({N}/{\log(N)}\bigg)^{-\frac{(1-\epsilon)\beta}{d}}[\log(N)]^{\Delta},
\end{eqnarray*}
 for some $\Delta>0$ depending on $\alpha, \beta, d$.
\end{lemma}
\begin{proof}
For fixed $(a, \epsilon)$, we define the deterministic integers
\begin{eqnarray}
	k^*=\ceil*{\frac{1}{2}n^{1-a}C_D^{-d} C_b^{-\frac{d}{\beta}}\delta^{\frac{d}{\beta}}}, \quad  k^\dagger=\ceil*{nC_D^d C_b^{-\frac{d}{\beta}}\delta^{\frac{d}{\beta}}},\nonumber
\end{eqnarray}
and events
\begin{eqnarray*}
E^*(a)=\bigg\{\|\bfX_{(k^*)}^j(\bfx)-\bfx\|\leq C_D\bigg(\frac{k^*}{n^{1-a}}\bigg)^{\frac{1}{d}}, \textrm{ for all } \bfx\in \Omega \textrm{ and } j=1,\ldots, m\bigg\}
\end{eqnarray*}
and
\begin{eqnarray*}
E^\dagger=\bigg\{\|\bfX_{(k^\dagger)}^j(\bfx)-\bfx\|\geq \frac{1}{C_D}\bigg(\frac{k^\dagger}{n}\bigg)^{\frac{1}{d}}, \textrm{ for all } \bfx\in \Omega \textrm{ and } j=1,\ldots, m\bigg\}.
\end{eqnarray*}
By  the definition of $v^*$, for any $(a, \epsilon)$, we can verify that $(1-\epsilon)(1-a)\geq v^*d/\beta$ and $n^{1-a}\delta^{\frac{d}{\beta}}=C_\delta^{\frac{d}{\beta}} n^{1-a}N^{-\frac{v^*d}{\beta}}[\log(N)]^{\frac{v^*d}{\beta}}= C_\delta^{\frac{d}{\beta}} N^{(1-\epsilon)(1-a)-\frac{v^*d}{\beta}}[\log(N)]^{\frac{v^*d}{\beta}}$ is diverging. Consequently, we may assume $k^*\geq 1$ and $k^\dagger \geq 1$. On event $E^*(a)$, it follows from the definition of $k_j^{\textrm{opt}}$ that
\begin{eqnarray}
\|\bfX_{(k^*)}^j(\bfx)-\bfx\|\leq C_D\bigg(\frac{k^*}{n^{1-a}}\bigg)^{\frac{1}{d}}&\leq& C_D\bigg(\frac{n^{1-a}C_D^{-d} C_b^{-\frac{d}{\beta}}\delta^{\frac{d}{\beta}}}{n^{1-a}}\bigg)^{\frac{1}{d}}\nonumber\\
&\leq& (C_b^{-1}\delta)^{\frac{1}{\beta}}<\|X_{(k_j^{\textrm{opt}}+1)}^j(\bfx)-\bfx\|.\nonumber
\end{eqnarray}
which further implies that
\begin{eqnarray}
	k_j^{\textrm{opt}}(\bfx)\geq k^*\geq \frac{1}{2}n^{1-a} C_D^{-d} C_b^{-\frac{d}{\beta}}\delta^{\frac{d}{\beta}} \label{eq:lemma:claim:3:eq:1}
\end{eqnarray}	
for all $\bfx\in \Omega \textrm{ and } j=1,\ldots, m.$ Moreover, on event $E^\dagger$, it also holds that
\begin{eqnarray}
\|\bfX_{(k_j^\dagger)}^j-\bfx\|\geq\frac{1}{C_D}\bigg(\frac{k^\dagger}{n}\bigg)^{\frac{1}{d}}\geq   (C_b^{-1}\delta)^{\frac{1}{\beta}}\geq \|X_{(k_j^{\textrm{opt}})}^j-\bfx\|,\nonumber
\end{eqnarray}
where the last equation follows from the definition of $k_j^{\textrm{opt}}$.  Above inequality  implies that
\begin{eqnarray}
	 k_j^{\textrm{opt}}(\bfx)\leq k^\dagger \leq n C_D^d C_b^{-\frac{d}{\beta}}\delta^{\frac{d}{\beta}}\label{eq:lemma:claim:3:eq:2}
\end{eqnarray}	
for all $\bfx \in\Omega \textrm{ and } j=1,\ldots, m.$
Combining (\ref{eq:lemma:claim:3:eq:1}) and (\ref{eq:lemma:claim:3:eq:2}), we conclude that on event $E^*(a)\cap E^\dagger$, the following holds:
\begin{eqnarray}
\frac{1}{2} n^{1-a} C_D^{-d} C_b^{-\frac{d}{\beta}}\delta^{\frac{d}{\beta}}\leq k_j^{\textrm{opt}}(\bfx)\leq n C_D^d C_b^{-\frac{d}{\beta}}\delta^{\frac{d}{\beta}} \quad \textrm{ for all } \bfx \in\Omega \textrm{ and } j=1,\ldots, m.\nonumber\\
\label{eq:lemma:claim:3:eq:3}
\end{eqnarray}

Define $k^{\min}=\min\{k_1^{\textrm{opt}}, k_2^{\textrm{opt}},\ldots, k_m^{\textrm{opt}}\}$. Hence, by (\ref{eq:lemma:claim:3:eq:3}), it holds on event $E^*(a)\cap E^\dagger$ that
\begin{eqnarray}
\frac{1}{2} n^{1-a} C_D^{-d} C_b^{-\frac{d}{\beta}}\delta^{\frac{d}{\beta}}\leq k^{\min}(\bfx)\leq n C_D^d C_b^{-\frac{d}{\beta}}\delta^{\frac{d}{\beta}} \quad \textrm{ for all } \bfx \in\Omega.\label{eq:lemma:claim:3:eq:4}
\end{eqnarray}
Since $k^{\min}(\bfx)\leq k_j^{\textrm{opt}}$, so under event $E^*(a)\cap E^\dagger$ , for all $\bfx$ with $\zeta(\bfx)\geq \delta$ and $f^*(\bfx)=1$, Lemma \ref{lemma:claim:1} and (\ref{eq:lemma:claim:3:eq:4}) together imply that
\begin{eqnarray}
&&	\sqrt{mk^{\min}}\bigg(\ev(\widehat{\eta}_{k^{\min}}(\bfx)|\mcX)-\frac{1}{2}\bigg)\nonumber
	\\&\geq& \nonumber \sqrt{\frac{1}{2} mn^{1-a} C_D^{-d} C_b^{-\frac{d}{\beta}}\delta^{\frac{d}{\beta}}} c_b\zeta(\bfx)\\
	&\geq& \sqrt{\frac{1}{2}c_b^2C_D^{-d} C_b^{-\frac{d}{\beta}}}\sqrt{m^aN^{1-a}\delta^{\frac{d}{\beta}+2}} \nonumber\\
	&=& \sqrt{\frac{1}{2}c_b^2C_D^{-d} C_b^{-\frac{d}{\beta}}C_\delta^{\frac{2\beta+d}{\beta}}}\sqrt{N^{\epsilon a}N^{1-a}\bigg({N}/{\log(N)}\bigg)^{-\frac{v^*(2\beta+d)}{\beta}}} \nonumber.
\end{eqnarray}
Since by definition, $v^*=(1-\epsilon)(1-a)\beta/d$ if $\epsilon\geq 2\beta/(2\beta+d), 0<a<1$, it holds that
\begin{eqnarray}
\frac{v^*(2\beta+d)}{\beta}=\frac{(1-\epsilon)(1-a)(2\beta+d)}{d}\leq 1-a.\nonumber
\end{eqnarray}
Combining the above two inequalities, we have
\begin{eqnarray}
	&&\sqrt{mk^{\min}}\bigg(\ev(\widehat{\eta}_{k^{\min}}(\bfx)|\mcX)-\frac{1}{2}\bigg)\nonumber\\
	&\geq&
	\begin{cases}
	 \sqrt{\frac{1}{2}c_b^2C_D^{-d} C_b^{-\frac{d}{\beta}}C_\delta^{\frac{2\beta+d}{\beta}}}\sqrt{N\bigg({N}/{\log(N)}\bigg)^{-1}}  &\textrm{ if } \epsilon<\frac{2\beta}{2\beta+d}, a=1\\
	 \sqrt{\frac{1}{2}c_b^2C_D^{-d} C_b^{-\frac{d}{\beta}}C_\delta^{\frac{2\beta+d}{\beta}}}\sqrt{N^{\epsilon a} N^{1-a}\bigg({N}/{\log(N)}\bigg)^{-(1-a)}}  &\textrm{ if } \epsilon\geq \frac{2\beta}{2\beta+d}, a>0\\
	\end{cases}\nonumber\\
	&\geq&3\sqrt{\frac{(d+2)\log(N)}{2}},\nonumber
\end{eqnarray}
where the last inequality follows if we choose $C_\delta$ large. By above inequality, on event $E_A\cap E^*(a)\cap E^\dagger$, for all $\bfx$ with $\zeta(\bfx)\geq \delta$ and $f^*(\bfx)=1$, it follows that
 \begin{eqnarray}
	&&\sqrt{mk^{\min}}\bigg(\widehat{\eta}_{k^{\min}}(\bfx)-\frac{1}{2}\bigg)\nonumber\\
	&\geq&\sqrt{mk^{\min}}\bigg(\ev(\widehat{\eta}_{k^{\min}}(\bfx)|\mcX)-\frac{1}{2}\bigg)-\sqrt{mk^{\min}}\bigg|\widehat{\eta}_{k^{\min}}(\bfx)-\ev(\widehat{\eta}_{k^{\min}}(\bfx)|\mcX)\bigg|\nonumber\\
	&\geq&3\sqrt{\frac{(d+2)\log(N)}{2}}-\sqrt{\frac{(d+2)\log(N)}{2}}\nonumber\\
	&>& \sqrt{\frac{(d+2)\log(N)}{2}}.\nonumber
\end{eqnarray}
Similarly, on event $E_A\cap E^*(a)\cap E^\dagger$, for all $\bfx$ with $\zeta(\bfx)\geq \delta$ and $f^*(\bfx)=0$, we can show that
\begin{align}
	\sqrt{mk^{\min}}\bigg(\widehat{\eta}_{k^{\min}}(\bfx)-\frac{1}{2}\bigg)<- \sqrt{\frac{(d+2)\log(N)}{2}}.\nonumber
\end{align}
Therefore, we prove that on event $E_A\cap E^*(a)\cap E^\dagger$, the following holds:
\begin{eqnarray}
		\sqrt{mk^{\min}}\bigg|\widehat{\eta}_{k^{\min}}(\bfx)-\frac{1}{2}\bigg|> \sqrt{\frac{(d+2)\log(N)}{2}} \textrm{ for all } \bfx \textrm{ with } \zeta(\bfx)\geq \delta.\label{eq:lemma:claim:3:eq:4.5}
\end{eqnarray}
Now by (\ref{eq:lemma:claim:3:eq:4}) and the definition of $v^*$, on event $E_A\cap E^*(a)\cap E^\dagger$, we have
\begin{eqnarray}
k^{\min}(\bfx)&\leq&  n C_D^{d} C_b^{-\frac{d}{\beta}}\delta^{\frac{d}{\beta}}\nonumber\\
&=&C_D^{d} C_b^{-\frac{d}{\beta}}C_\delta^{\frac{d}{\beta}} n \bigg({N}/{\log(N)}\bigg)^{-\frac{v^*d}{\beta}}\nonumber\\
& \leq&   C_D^{-d} C_b^{-\frac{d}{\beta}}C_\delta^{\frac{d}{\beta}} nN^{-(1-\epsilon)(1-a)}\log(N) \quad \textrm{ if } \epsilon\geq \frac{2\beta}{2\beta+d},  0<a<1,\nonumber\\\label{eq:lemma:claim:3:eq:5}
\end{eqnarray}
for all $\bfx$ with $\zeta(\bfx)\geq \delta$.  In the following, we will calculate the probability of $E_A\cap E^*(a)\cap E^\dagger$. Since $\alpha \beta\leq d$ by Assumption \ref{A1:marginal:assumption}, it follows that
\begin{eqnarray*}
v^*(1+\alpha)=\frac{(1-\epsilon)(1-a)\beta(1+\alpha)}{d}\leq \frac{(1-a)\beta(1+\alpha)}{2\beta+d}&\leq& \frac{\beta(1+\alpha)}{2\beta+d}\\
&=& \frac{\beta+\alpha\beta}{2\beta+d}\leq \frac{\beta+d}{2\beta+d}< 1.
\end{eqnarray*}
Using the inequality above and (\ref{eq:probability:EA}), it follows that
\begin{eqnarray}
	\pr(E_A)\geq 1-dN^{-1}\geq 1-dC_\delta^{1+\alpha}N^{-v^*(1+\alpha)}[\log(N)]^{v^*(1+\alpha)}=1-d\delta^{1+\alpha},\nonumber	
\end{eqnarray}
for sufficiently large $N$.  Moreover, by Lemma \ref{lemma:order:statistics:bound} and the definition of $E^*(a), E^\dagger$, it follows that 
\begin{eqnarray}
	\pr(E^*(a))&\geq& 1-C_D\frac{mn^{1-a}}{k^*}\exp(-n^ak^*/6)\nonumber\\
	&\geq&1-C_DN\exp\bigg(-\frac{n\delta^{\frac{d}{\beta}}}{12C_D^{d} C_b^{\frac{d}{\beta}}}\bigg)\nonumber\\
		&=&1-C_DN\exp\bigg\{-\frac{1}{12C_D^{d} C_b^{\frac{d}{\beta}}}N^{1-\epsilon}\bigg({N}/{\log(N)}\bigg)^{-\frac{v^*d}{\beta}}\bigg\}\nonumber\\
		&=&1-C_DN\exp\bigg\{-\frac{1}{12C_D^{d} C_b^{\frac{d}{\beta}}}N^{(1-\epsilon)a}[\log(N)]^{(1-\epsilon)(1-a)}\bigg\}\nonumber
\end{eqnarray}
and
\begin{eqnarray*}
\pr(E^\dagger)&=&1-C_D\frac{mn}{k^\dagger}\exp(-k^\dagger/6)\nonumber\\
&\geq& 1-C_DN \exp\bigg(-\frac{1}{12C_D^{-d} C_b^{\frac{d}{\beta}}} n \delta^{\frac{d}{\beta}}\bigg)\\
&\geq&1-C_DN\exp\bigg\{-\frac{1}{12C_D^{-d} C_b^{\frac{d}{\beta}}}N^{(1-\epsilon)a}[\log(N)]^{(1-\epsilon)(1-a)}\bigg\}\nonumber,
\end{eqnarray*}
Combining the above three inequalities, we show that 
\begin{eqnarray}
\pr(E_A\cap E^*(a) \cap E^\dagger)&\geq& 1-\pr(E_A)-\pr(E^*(a))-\pr(E^\dagger)\nonumber\\
&\geq& 1-d\delta^{1+\alpha}-CN\exp\bigg\{-\frac{1}{C}N^{(1-\epsilon)a}[\log(N)]^{(1-\epsilon)(1-a)}\bigg\} \nonumber\\
&\geq& 1-d\delta^{1+\alpha}-CN\exp\bigg\{-\frac{1}{C}N^{(1-\epsilon)a}\bigg\} \nonumber,
\end{eqnarray}
where $C$ is some constant greater than $C_D+12C_D^dC_b^{\frac{d}{\beta}}+12C_D^{-d}C_b^{\frac{d}{\beta}}$. Without loss of generality, we assume $C>6$. if we choose $a=\frac{\log(2C\log(N))}{(1-\epsilon)\log(N)}$, then we have
\begin{eqnarray}
N^{a}&=&[2C\log(N)]^{\frac{1}{1-\epsilon}}\nonumber\\
\delta&=&C_\delta\bigg({N}/{\log(N)}\bigg)^{-\frac{(1-\epsilon)(1-a)\beta}{d}}\leq C_\delta\bigg({N}/{\log(N)}\bigg)^{-\frac{(1-\epsilon)\beta}{d}}[2C\log(N)]^{\frac{\beta}{d}},\nonumber\\
\label{eq:lemma:claim:3:eq:5.6}
\end{eqnarray}
and the probability can be bounded by
\begin{eqnarray}
\pr(E_A\cap E^*(a) \cap E^\dagger)\geq 1-\delta^{1+\alpha}-CN^{-1}\geq 1-(C+1)\delta^{1+\alpha}.\label{eq:lemma:claim:3:eq:5.5}
\end{eqnarray}

First, let us consider the case $\epsilon<\frac{2}{2+d}$. By  (\ref{eq:lemma:claim:3:eq:4.5}), (\ref{eq:lemma:claim:3:eq:5}) and the definition of $\widehat{k}(\bfx)$, since $\frac{2\beta}{2\beta+d}\leq \epsilon< \frac{2}{2+d}$ and $0<a=\frac{\log(2C\log(N))}{(1-\epsilon)\log(N)}\leq 1-\frac{d}{(2+d)(1-\epsilon)}$ for large $N$,  we have
\begin{eqnarray*}
 \widehat{k}(\bfx)\leq k^{\min}(\bfx)\leq nN^{-\frac{d}{2+d}}\log(N) \quad \textrm{ and } \quad \widehat{k}(\bfx)\leq k^{\min}(\bfx)\leq k_j^{opt}(\bfx)
\end{eqnarray*}
holds for all $j=1,\ldots, m$ and all $\bfx$ with $\zeta(\bfx)\geq \delta$ on event $E_A\cap E^*(a)\cap E^\dagger$.
Now, applying  Lemma \ref{lemma:claim:2}, we can see, it holds on event $E_A\cap E^*(a)\cap E^\dagger$ that: 
\begin{eqnarray}
	\widehat{f}_{\widehat{k}}(\bfx)=f^*(\bfx), \textrm{ for all } \bfx \textrm{ with } \zeta(\bfx)\geq \delta.\nonumber
\end{eqnarray}
By (\ref{eq:lemma:claim:3:eq:5.5}), (\ref{eq:lemma:claim:3:eq:5.6}) and the above equation, we can complete the proof using the same argument as  (\ref{eq:lemma:claim:3:eq:8:unequal}).

Second, let us assume $\epsilon=\frac{2}{2+d}\geq \frac{2\beta}{2\beta+d}$. By (\ref{eq:lemma:claim:3:eq:5})  and the definition of $\widehat{k}(\bfx)$,  the following holds on event $E_A\cap E^*(a) \cap E^\dagger$:
\begin{eqnarray*}
\widehat{k}(\bfx)\leq k^{\min}(\bfx)\leq  C_D^{-d} C_b^{-\frac{d}{\beta}}C_\delta^{\frac{d}{\beta}} nN^{-(1-\epsilon)(1-a)}\log(N)= C_D^{-d} C_b^{-\frac{d}{\beta}}C_\delta^{\frac{d}{\beta}} N^{(1-\epsilon)a}\log(N):=v_N.
\end{eqnarray*}
for all $\bfx$ with $\zeta(\bfx)\geq \delta$, which further leads to
\begin{eqnarray}
C_b N^{-\frac{(1-\epsilon)(1-a)\beta}{d}}\leq C_b[\widehat{k}(\bfx)/N^{(1-\epsilon)(1-a)}]^{\frac{\beta}{d}} \leq C_b v_N^{\frac{\beta}{d}}N^{-\frac{(1-\epsilon)(1-a)\beta}{d}}. \label{eq:lemma:claim:3:eq:7}
\end{eqnarray}


Let us define classifier 
\begin{eqnarray*}
\widetilde{f}(\bfx)=\begin{cases}
f^*(\bfx) & \textrm{ if }  \widehat{f}_k(\bfx)=f^*(\bfx) \textrm{ for all } 1\leq k \leq v_N;\\
1-f^*(\bfx) & \textrm{ elsewhere}.
\end{cases}
\end{eqnarray*}
By the definition of $\widetilde{f}(\bfx)$, it follows that
\begin{eqnarray*}
\pr(\widetilde{f}(\bfx)\neq f^*(\bfx)|\mcX)&\leq& \pr(\exists 1\leq k \leq v_N, \textrm{ such that } \widehat{f}_k(\bfx)\neq f^*(\bfx)|\mcX)\\
&\leq& \sum_{k=1}^{\floor{v_N}}\pr(\widehat{f}_k(\bfx)\neq f^*(\bfx)|\mcX).
\end{eqnarray*}
By the above inequality, (\ref{eq:lemma:claim:3:eq:7}) and Lemma \ref{lemma:case:1:exponential:inequality}, we conclude the following hold on event $\cap_{k=1}^{\floor{v_N}}E_P(k:k, a)\cap E_A\cap E^*(a)\cap E^\dagger$:
\begin{eqnarray}
\pr(\widehat{f}_{\widehat{k}}(\bfx)\neq f^*(\bfx)|\mcX)=\pr(\widehat{f}_{\widehat{k}}(\bfx)\neq f^*(\bfx)|\mcX)&\leq&  \pr(\widetilde{f}(\bfx)\neq f^*(\bfx)|\mcX)\nonumber\\
&\leq& v_N\exp\bigg(-{2c_b^2 m \zeta^{2}(\bfx)}\bigg)\nonumber\\\label{eq:lemma:claim:3:eq:8},
\end{eqnarray}
for all $\bfx$ with $\zeta(\bfx)\geq \max\{\delta, C_bv_N^{\frac{\beta}{d}}N^{-\frac{(1-\epsilon)(1-a)\beta}{d}}\}$. For simplicity, let us denote $\widetilde{\delta}=\max\{\delta, C_bv_N^{\frac{\beta}{d}}N^{-\frac{(1-\epsilon)(1-a)\beta}{d}}\}$, $E=\cap_{k=1}^{\floor{v_N}}E_P(k:k, a)\cap E_A\cap E^*(a)\cap E^\dagger$, $A_0=\{\bfx: \|\eta(\bfx)-1/2\|\leq \widetilde{\delta}\}$, and $A_j=\{\bfx: 2^{j-1}<\|\eta(\bfx)-1/2\|\leq 2^j\widetilde{\delta}\}$. 

If $j=0$, then Assumption \ref{A1:marginal:assumption} shows that
\begin{eqnarray}
	\ev\bigg(|2\eta(\bfX)-1|\mbI(\widehat{f}_{\widehat{k}}(\bfX)\neq f^*(\bfX))\mbI(\bfX\in A_0, E)\bigg)\leq2 \widetilde{\delta} P(\bfX\in A_0)\leq 2C_\alpha \widetilde{\delta}^{1+\alpha}.\nonumber
\end{eqnarray}

If $j\geq 1$,  (\ref{eq:lemma:claim:3:eq:8}) implies that
\begin{eqnarray}
	&&\ev\bigg(|2\eta(\bfX)-1|\mbI(\widehat{f}_{\widehat{k}}(\bfX)\neq f^*(\bfX))\mbI(\bfX\in A_j, E)\bigg)\nonumber\\
	&\leq&2^{j+1} \widetilde{\delta} \ev\bigg\{\mbI(\bfX\in A_j,  E)\pr(\widehat{f}_{\widehat{k}}(\bfX)\neq f^*(\bfX) | \bfX, \mcX)\bigg\}\nonumber\\
	&\leq& 2^{j+1} \widetilde{\delta} v_N\ev\bigg\{\mbI(\bfX\in A_j, E)\exp\bigg(-{2c_b^2 m\zeta^{2}(\bfx)}\bigg)\bigg\}\nonumber\\
		&\leq& 2^{j+1} \widetilde{\delta} v_N\ev\bigg\{\mbI(\bfX\in A_j, E)\exp\bigg(-{2c_b^2m 4^{j-1}\widetilde{\delta}^{2}}\bigg)\bigg\}.\nonumber
\end{eqnarray}

Since $\widetilde{\delta}=\max\{\delta, C_bv_N^{\frac{\beta}{d}}N^{-\frac{(1-\epsilon)(1-a)\beta}{d}}\}= C_D^dC_b^{\frac{d+\beta}{\beta}}C_\delta^{\frac{d}{\beta}}N^{\frac{(d+\beta)a-\beta}{2+d}}\log(N)$ and $m=N^{\epsilon}=N^{\frac{2}{2+d}}$, so $m\widetilde{\delta}^2\geq 1$ for all $a>0$. As a consequence of Assumption \ref{A1:marginal:assumption}, it follows that
\begin{eqnarray*}
&&\sum_{j=1}^\infty \ev\bigg(|2\eta(\bfX)-1|\mbI(\widehat{f}_{\widehat{k}}(\bfX)\neq f^*(\bfX))\mbI(\bfX\in A_j, E)\bigg)\nonumber\\
&\leq& \widetilde{\delta} v_N \sum_{j=1}^\infty 2^{j+1} \exp\bigg(-{2c_b^2 4^{j-1}}\bigg) \pr(A_j)\\
&\leq& 2\widetilde{\delta}^{1+\alpha} v_N \sum_{j=1}^\infty 2^{j(1+\alpha)} \exp\bigg(-{\frac{1}{2}c_b^2 4^{j}}\bigg)\leq \widetilde{C} \widetilde{\delta}^{1+\alpha}v_N,
\end{eqnarray*}
where we use the fact that  $\sum_{j=1}^\infty b^j\exp(-c4^j)<\infty$ for all $b, c>0$, and $\widetilde{C}>0$ is a constant free of $N$. Combining the bounds above with (\ref{eq:probability:Ep}) and (\ref{eq:lemma:claim:3:eq:5.5}), it follows that
\begin{eqnarray}
	\mcR(\widehat{f}_{\widehat{k}})&=&\ev\big(|2\eta(\bfX)-1|\mbI(\widehat{f}_{\widehat{k}}(\bfX)\neq f^*(\bfX))\big)\nonumber\\
	&\leq&\ev\big(|2\eta(\bfX)-1|\mbI(\widehat{f}_{\widehat{k}}(\bfX)\neq f^*(\bfX), E^c)\big)\nonumber\\
	&&+\ev\big(|2\eta(\bfX)-1|\mbI(\widehat{f}_{\widehat{k}}(\bfX)\neq f^*(\bfX), \bfX\in A_0,E_P)\big)\nonumber\\
	&&+\sum_{j=1}^\infty\ev\big(|2\eta(\bfX)-1|\mbI(\widehat{f}_{\widehat{k}}(\bfX)\neq f^*(\bfX), \bfX\in A_j,E)\big)\nonumber\\
&\leq& \pr(E^c)+\widetilde{C}\widetilde{\delta}^{1+\alpha}v_N\nonumber\\
&\leq& \sum_{k=1}^{\floor{v_N}}\pr(E_P^c(k,a))+\pr((E_A^c\cap E^*(a)\cap E^\dagger)^c)+\widetilde{C}\widetilde{\delta}^{1+\alpha}v_N\nonumber\\
&\leq& C_Dv_NN\exp(-n^a/6)+(1+C)\delta^{1+\alpha}+\widetilde{C}\widetilde{\delta}^{1+\alpha}v_N\nonumber.
\end{eqnarray}
Notice that $a=\frac{\log(2C\log(N))}{(1-\epsilon)\log(N)}$ with $C>6$, we have
\begin{eqnarray*}
v_N&=&C_D^{-d} C_b^{-\frac{d}{\beta}}C_\delta^{\frac{d}{\beta}} N^{(1-\epsilon)a}\log(N)=2CC_D^{-d} C_b^{-\frac{d}{\beta}}C_\delta^{\frac{d}{\beta}} [\log(N)]^2,\\
\delta&\leq& \widetilde{\delta}= C_bv_N^{\frac{\beta}{d}}N^{-\frac{(1-\epsilon)(1-a)\beta}{d}}\lesssim N^{-\frac{(1-\epsilon)\beta}{d}}[\log(N)]^{\frac{3\beta}{d}},\\
\exp(-n^a/6)&=&\exp(-[2C\log(N)]/6)\leq N^{-2}.
\end{eqnarray*}
Since $\epsilon=\frac{2}{2+d}$, we conclude that
\begin{eqnarray*}
\mcR(\widehat{f}_{\widehat{k}})\lesssim  \bigg({N}/{\log(N)}\bigg)^{-\frac{(1-\epsilon)\beta(1+\alpha)}{d}}[\log(N)]^{\Delta} \textrm{ for some } \Delta>0.
\end{eqnarray*}
\end{proof}

\begin{lemma}\label{lemma:claim:4}
Under Assumptions \ref{A1:strong:density}-\ref{A1:marginal:assumption}, if $\epsilon\geq 2\beta/(2\beta+d)$ and $k=1$, then   $$\mcR(\widehat{f}_{k})\lesssim  N^{-\frac{(1-\epsilon)\beta(1+\alpha)}{d}}[\log(N)]^\Delta$$ for some $\Delta>0$ depending on $\alpha, \beta, d$ and $\epsilon$.
\end{lemma}
\begin{proof}
Let us define ${\widetilde{\delta}}= C_b[k/N^{(1-\epsilon)(1-a)}]^{\frac{\beta}{d}}$, where $C_b$ is the constant in Lemma \ref{lemma:bias:bound}, and we allow $a$ is depending on $N$. 
We further define  sets $A_0=\{\bfx : |\eta(\bfx)-1/2|\leq {\widetilde{\delta}}\}$ and $A_j=\{\bfx : 2^{j-1}{\widetilde{\delta}}<|\eta(\bfx)-1/2|\leq 2^j{\widetilde{\delta}}\}$ for $j\geq 1$. For simplicity, we write $E_P(k_1:k_m, a)=E_P$ and in the equal sub-sample size setting, we have $k_1=\ldots=k_m=k=1$.

If $j=0$, then Assumption \ref{A1:marginal:assumption} shows that
\begin{eqnarray}
	\ev\bigg(|2\eta(\bfX)-1|\mbI(\widehat{f}_{k}(\bfX)\neq f^*(\bfX))\mbI(\bfX\in A_0, E_P)\bigg)\leq2\delta \pr(\bfX\in A_0)\leq 2C_\alpha {\widetilde{\delta}}^{1+\alpha}.\nonumber
\end{eqnarray}

If $j\geq 1$, Assumption \ref{A1:marginal:assumption} and Lemma \ref{lemma:case:1:exponential:inequality} imply that
\begin{eqnarray}
	&&\ev\bigg(|2\eta(\bfX)-1|\mbI(\widehat{f}_{k}(\bfX)\neq f^*(\bfX))\mbI(\bfX\in A_j, E_P)\bigg)\nonumber\\
	&\leq&2^{j+1}\delta \ev\bigg\{\mbI(\bfX\in A_j,  E_P)\pr(\widehat{f}_{k}(\bfX)\neq f^*(\bfX) | \bfX, \mcX)\bigg\}\nonumber\\
	&\leq& 2^{j+1}\delta \ev\bigg\{\mbI(\bfX\in A_j, E_P)\exp\bigg(-{2c_b^2 mk\zeta^{2}(\bfx)}\bigg)\bigg\}\nonumber\\
		&\leq& 2^{j+1}\delta \ev\bigg\{\mbI(\bfX\in A_j, E_P)\exp\bigg(-{2c_b^2m4^{j-1}{\widetilde{\delta}}^{2}}\bigg)\bigg\}\label{eq:theorem:average:estimator:eq046}.
\end{eqnarray}
Since ${\widetilde{\delta}}=C_b [k/N^{(1-\epsilon)(1-a)}]^{\frac{\beta}{d}}$ and $\epsilon\geq 2\beta/(2\beta+d)$,  any $a\in (0, 1)$ satisfies $1-a\leq \frac{d\log( m)}{2\beta(1-\epsilon)\log(N)}=\frac{d\epsilon}{2\beta (1-\epsilon)}$. Therefore, we can pick any $a\in (0,1)$ and it follows that $m\geq N^{\frac{2\beta (1-\epsilon)(1-a)}{d}}$. By the choice with  $k=1$,  we have ${\widetilde{\delta}}=C_b N^{-\frac{\beta (1-\epsilon) (1-a)}{d}}$ and
\begin{eqnarray}
(\ref{eq:theorem:average:estimator:eq046})&\leq& 2^{j+1}{\widetilde{\delta}} \ev\bigg\{\mbI(\bfX\in A_j, E_P)\exp\bigg(-{\frac{1}{2}c_b^2 C_b^2 4^{j} m [N^{-(1-\epsilon)(1-a)}]^{\frac{2\beta}{d}}}\bigg)\bigg\}\nonumber\\
&\leq& 2^{j+1}{\widetilde{\delta}} \ev\bigg\{\mbI(\bfX\in A_j, E_P)\exp\bigg(-{\frac{1}{2}c_b^2 C_b^2 4^{j}}\bigg)\bigg\}\nonumber\\
&=&  2C_\alpha {\widetilde{\delta}}^{1+\alpha} 2^{j(1+\alpha)}\exp\bigg(-{\frac{1}{2}c_b^2 C_b^2 4^{j}}\bigg).\nonumber
\end{eqnarray}
Taking summation ans using the fact that $\sum_{j=1}^\infty b^j \exp(-c4^j)<\infty$ for all $b,c>0$, we have
\begin{eqnarray}
\sum_{j=1}^\infty\ev\bigg(|2\eta(\bfX)-1|\mbI(\widehat{f}_{k}(\bfX)\neq f^*(\bfX))\mbI(\bfX\in A_j, E_P)\bigg)\leq C{\widetilde{\delta}}^{1+\alpha},\nonumber
\end{eqnarray}
where $C$ is some constant free of $N$ and $a$.
Combining the bounds above, it follows that
\begin{eqnarray}
	\mcR(\widehat{f}_{k})&=&\ev\big(|2\eta(\bfX)-1|\mbI(\widehat{f}_{k}(\bfX)\neq f^*(\bfX))\big)\nonumber\\
	&\leq&\ev\big(|2\eta(\bfX)-1|\mbI(\widehat{f}_{k}(\bfX)\neq f^*(\bfX), E_P^c)\big)\nonumber\\
	&&+\ev\big(|2\eta(\bfX)-1|\mbI(\widehat{f}_{k}(\bfX)\neq f^*(\bfX), \bfX\in A_0,E_P)\big)\nonumber\\
	&&+\sum_{j=1}^\infty\ev\big(|2\eta(\bfX)-1|\mbI(\widehat{f}_{k}(\bfX)\neq f^*(\bfX), \bfX\in A_j,E_P)\big)\nonumber\\
&\leq& \pr(E_P^c)+C{\widetilde{\delta}}^{1+\alpha}\nonumber\\
&=&\pr(E_P^c)+CC_b^{1+\alpha}N^{-\frac{(1-\epsilon)(1-a)(1+\alpha)\beta}{d}}.\nonumber
\end{eqnarray}
Let $a=\frac{s\log(\log(N))}{\log(N)}$ for some $s>0$ and notice that  $N^{\frac{s\log(\log(N))}{\log(N)}}=e^{s\log(\log(N))}=[\log(N)]^s$. As a consequence, it follows that
\begin{eqnarray*}
N^{-\frac{(1-\epsilon)(1-a)(1+\alpha)\beta}{d}}=N^{-\frac{(1-\epsilon)(1+\alpha)\beta}{d}}[\log(N)]^{\frac{s(1-\epsilon)(1+\alpha)\beta}{d}}.
\end{eqnarray*}
By (\ref{eq:probability:Ep}), since $\epsilon \geq 2\beta/(2\beta+d)$ and $k=1$, if we choose $s$ such that $2^{s(1-\epsilon)}= 48$, then
\begin{eqnarray*}
\pr(E_P^c)&\leq& C_Dmn^{1-a} \exp({-n^a/6})\\
&\leq& C_DN\exp({-n^a/6})\\
&=&C_DN\exp(-N^{(1-\epsilon)a}/6)\\
&=&C_DN\exp\bigg(-[\log(N)]^{s(1-\epsilon)}/6\bigg)\\
&=& C_DN\exp\bigg(-\log(N)[\log(N)]^{s(1-\epsilon)-1}/6\bigg)\\
&\leq&C_DN\exp\bigg(-\log(N)2^{s(1-\epsilon)-1}/6\bigg)\\
&\leq&C_DN\exp\bigg(-4\log(N)\bigg)\leq C_DN^{-3}.
\end{eqnarray*}
Combining the above three inequalities and noticing that $\frac{(1-\epsilon)(1+\alpha)\beta}{d}\leq \frac{(1+\alpha)\beta}{d}=\frac{1+\alpha\beta}{d}\leq \frac{1+d}{d}\leq 2$ by Assumption \ref{A1:marginal:assumption}, we complete the proof with $\Delta=\frac{s(1-\epsilon)(1+\alpha)\beta}{d}$ and $s=\frac{\log(48)}{(1-\epsilon)\log(2)}$.
\end{proof}

Based on the above lemmas, we are ready to prove Theorem  \ref{theorem:adaptive:estimator}.

If $\epsilon<\frac{2\beta}{2\beta+d}$, it follows from Theorem \ref{theorem:adaptive:estimator:unequal}.

If $\epsilon>\frac{2}{2+d}$, then $\epsilon\geq \frac{2\beta}{2\beta+d}$ for any $\beta\in (0, 1]$. As a consequence, we have $nN^{-\frac{d}{2+d}}\log(N)=N^{1-\epsilon-\frac{d}{2+d}}\log(N)<1$ and $\widehat{k}=1$. The desired result follows from Lemma \ref{lemma:claim:4}.

If $\frac{2\beta}{2\beta+d}\leq \epsilon \leq \frac{2}{2+d}$, the convergence rate follows from Lemma \ref{lemma:claim:3}.

\subsection{Proof of Theorem \ref{theorem:1nn} }
Theorem \ref{theorem:1nn} follows directly from Lemma \ref{lemma:claim:4}.




\end{appendices}

\clearpage
\bibliography{ref}

\end{document}